\documentclass[12pt]{article}

\usepackage[numbers]{natbib}
\usepackage{scicite}

\usepackage{hyperref}
\usepackage{breakurl}
\usepackage{makecell}
\usepackage{amsmath}
\usepackage{graphicx}
\usepackage{epstopdf} 
\usepackage{multirow}
\usepackage{times}
\usepackage{latexsym}
\usepackage{amssymb}
\usepackage{subfigure}
\usepackage{graphicx}
\usepackage[T1]{fontenc}
\usepackage{hyperref} 
\usepackage{bm}
\usepackage{hyperref}
\usepackage{booktabs} 
\usepackage{colortbl}
\usepackage{microtype}
\usepackage[lined,boxed,commentsnumbered]{algorithm2e}
\usepackage{inconsolata}
\usepackage{xcolor}
\usepackage[final]{changes}    %final
\newenvironment{sciabstract}{%
\begin{quote} \bf}
{\end{quote}}

\newcounter{lastnote}

\title{Benchmarking Retrieval-Augmented Large Language Models in Biomedical NLP: Application, Robustness, and Self-Awareness}

\author
{Mingchen Li$^{1}$, Zaifu Zhan$^{3}$, Han Yang$^{2}$  Yongkang Xiao$^{2}$ \\Huixue Zhou$^{2}$, Jiatan Huang$^{1}$,Rui Zhang$^{1\ast}$\\
\\
\normalsize{$^{1}$Division of Computational Health Sciences, Department of Surgery}\\
\normalsize{ University of Minnesota,
Minneapolis, MN, USA}\\
\normalsize{$^{2}$Institute for Health Informatics}\\
\normalsize{University of Minnesota, Minneapolis, MN, USA}\\
\normalsize{$^{3}$ Department of Electrical and Computer Engineering}\\
\normalsize{University of Minnesota, Minneapolis, MN, USA }\\
\normalsize{$^\ast$Corresponding author: Rui Zhang, Ph.D; E-mail: ruizhang@umn.edu}
}

\date{}

\begin{document}

\baselineskip24pt


\maketitle

\begin{sciabstract}
  Abstract: To reduce hallucinations in large language models (LLMs), retrieval-augmented LLMs (RALs) retrieve supporting knowledge from external databases. However, their performance on biomedical natural language processing (NLP) tasks remains underexplored. We introduce BioRAB, a comprehensive evaluation framework assessing RALs across five biomedical NLP tasks and 11 datasets, using four testbeds: unlabeled robustness, counterfactual robustness, diverse robustness, and self-awareness. To improve RALs’ robustness and negative awareness, we propose a Detect-and-Correct strategy and a contrastive learning approach. Experimental results show that RALs generally outperform standard LLMs on most biomedical tasks, but still struggle with robustness and self-awareness, particularly under counterfactual and diverse scenarios. Our proposed methods significantly improve performance in robustness to unlabeled and counterfactual data, and increase the model’s ability to detect and avoid incorrect predictions. These findings highlight key limitations in current RALs and underscore the need for continued refinement to ensure reliability and accuracy in high-stakes biomedical applications.
 
 Teaser:  BioRAB evaluates retrieval-augmented language models on robustness and self-awareness in biomedical NLP tasks.

% \deleted{This paper introduces BioRAB to assess the robustness and self-awareness capabilities of RAL in the biomedical domain, and a Detect-and-Correct strategy and a contrastive learning approach. to help improve the obustness and neg-
% ative awarenesacross five biomedical NLP tasks spanning nine datasets. Our findings indicate that RALs continue to encounter challenges related to robustness and self-awareness. }

\end{sciabstract}
\maketitle

\section{Introduction}
Recently, significant progress has been made in large language models (LLMs). To adapt the LLM to the biomedical domain, several biomedical focused LLMs have been developed, such as MedLLaMA-13B~\citep{wu2024pmcllama} and Med-PaLM 2~\citep{singhal2025toward}. Despite demonstrating impressive general capabilities~\citep{li2024rt}, these models still face significant challenges, including factual hallucination ~\citep{ji2023survey}
, and absence of newly uploaded knowledge~\citep{ovadia2023fine}.

Retrieval-augmented language models (RALs)~\citep{li2024rt,lewis2020rag,li2023understand,huang2024ritek}, in contrast, can retrieve knowledge from an external datastore when needed, potentially reducing hallucination and improve the new knowledge adaption ability. The most common method is to use the designed retriever to retrieve the knowledge that is relevant to the input sentence, subsequently, the retrieved knowledge, along with the input sentence, is fed into the LLM to assist in generating the expected output.

In the question answering (QA) task, a retrieval-augmented language model can access knowledge from an unlabeled corpus, which serves as a retrievable knowledge base, such as PubMed.  The QA format allows the unlabeled corpus to potentially furnish answers to questions. However, for tasks like triple extraction, incorporating the unlabeled corpus may yield adverse effects. Counterfactual information, such as error annotations, is prevalent in labeled corpora, presenting challenges for retrievers in obtaining useful information. Additionally, LLMs still grapple with generating unreliable information retrieved incorrectly. 
Incorporating diverse knowledge holds promise for improving model performance. For example, question answering relies on extracting information from extensive contexts, thus potentially impacting information extraction performance. Moreover, the influence of retrieval information from various tasks or datasets on RAL performance remains underexplored. Self-awareness is crucial for RALs, 
if RALs can distinguish between positive retrieval knowledge and negative knowledge, they have the opportunity to rectify their actions.

\begin{figure}[t]
        \centering
        \includegraphics[width=0.8\columnwidth]{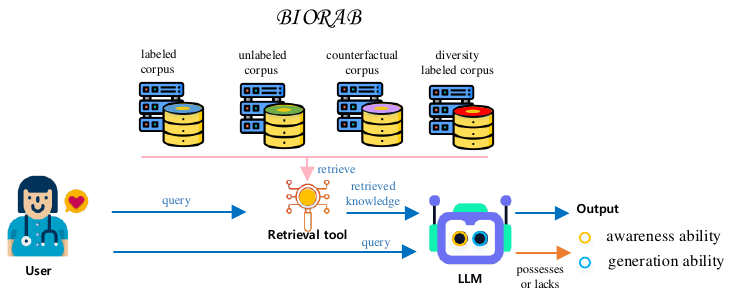}
	\caption{ BIORAB features on queries on different types corpus to test the awareness ability and generation ability of RAL.} 
	\label{con:whol_flow}
\end{figure}

These challenges hinder RALs from consistently producing reliable and accurate responses. Unfortunately, in the biomedical domain, only a few studies, such as Almanac~\citep{zakka2024almanac}, have explored the RAL performance in QA, leaving a gap in understanding how these factors affect RAL across various biomedical NLP tasks. Consequently, there is a pressing need for a comprehensive evaluation of RALs with different LLMs across biomedical NLP tasks.
%%%%%%%%%%%
To this end, this paper conducts a comprehensive evaluation of RAL for different LLMs on 5 biomedical NLP tasks over  11 datasets. Specifically, we create a new RAL benchmark for biomedical NLP, namely  BioRAB, as shown in Figure~\ref{con:whol_flow}, and create 4 testbeds to evaluate the mentioned fundamental abilities.

\begin{enumerate}
     \item\textbf{Unlabeled Robustness} denotes the ability of RALs to extract valuable information from unlabeled retrieval corpus, especially on label-intensive tasks, such as triple extraction, and classification.
     %%%%%%%%%%%
     For instance, in tasks like relation extraction, the corpus could be a labeled dataset (such as the training dataset) or unlabeled (training dataset without labels).
    If the RAL  achieves comparable or superior performance by retrieving the unlabeled dataset compared to retrieving the labeled dataset, it indicates that labeled databases may not be necessary for RALs. In the testbed of the Unlabeled Robustness, the corpus contains instances without labels.
     \item \textbf{Counterfactual Robustness} denotes whether the RAL could retrieve the right information from the counterfactual corpus, in our work, the counterfactual instance refers to the mislabeled annotation. In the testbed of Counterfactual Robustness,  the corpus consists of instances with a certain proportion of incorrect labels.
      \item \textbf{Diverse Robustness} evaluating whether RALs can achieve better performance by integrating information from multiple tasks.  For instance, the corpus for the classification task is sourced from relation extraction and question-answering tasks.  In the testbed of Diverse Robustness,  the corpus comprises instances from various tasks.  
     \item \textbf{Negative Awareness} refers to the RAL's ability to discern whether retrieved knowledge positively or negatively impacts the final output.  In the testbed of Negative Awareness,  the corpus comprises instances that are  100\% counterfactual instances.

\end{enumerate}

Utilizing BioRAB, we evaluate its performance across 5 tasks (triple extraction, link prediction, text classification, question answering, and natural language inference) using 11 biomedical NLP datasets. Furthermore, BioRAB undergoes evaluation with five widely used LLMs: LLaMA2-13B~\citep{touvron2023llama} , MedLLaMA-13B~\citep{wu2024pmcllama}, LLaMA3-8B~\citep{touvron2024llama3}, Phi4 14B~\citep{abdin2024phi4} and Qwen2.5 32B~\citep{yang2024qwen2}, utilizing three commonly employed retrievers (BM25~\citep{robertson1976relevance}, Contriever~\citep{izacard2021unsupervised}, and MedCPT~\citep{jin2023medcpt}.

We observed that although RALs can enhance response accuracy in the majority of biomedical NLP tasks we evaluated, they encounter notable challenges. Particularly in the question-answering task, we noted that RALs did not yield significant improvements in the datasets we used. We speculate that this could be attributed to the limitations of the corpus used for retrieving, as the training dataset (corpus we used for retrieving) may not have provided sufficient information compared to using Wikipedia or PubMed.  Moreover, RALs struggle to generate the desired output when the corpus lacks labeling when compared to the labeled corpus. An interesting finding is that in datasets like ChemProt and Hetionet, RALs exhibit improved performance with unlabeled corpora compared to the source LLM.  Besides, RALs lack the capability to extract useful information from counterfactual corpora and struggle to discern the most relevant information. We also find this is not a common case, some datasets, such as in the dataset ADE and Hetionet, RAL could handle the counterfactual instance. Additionally, when presented with a diverse labeled corpus, RALs do not achieve optimal performance across tasks, except for the natural language inference task. Finally, we found despite providing counterfactual examples during training, the LLM was still able to generate correct outputs in some instances. However, RALs struggle with self-awareness, as they lack the ability to determine which examples could help improve model performance. 
%%%
% The experimental results mentioned above underscore the necessity for further resolution of these issues for the RAL.

To address these challenges, we propose a Detect-and-Correct method aimed at enhancing RAL’s ability to identify and rectify errors in the retrieved corpus, thereby improving its overall reliability. Additionally, we introduce a contrastive learning approach that refines RAL’s negative awareness by strengthening its capacity to distinguish differences of samples. By integrating these two techniques, our method significantly enhances RAL’s robustness and awareness ablilty, making it more effective in processing complex retrieval tasks. Together, these improvements contribute to a more resilient and accurate model for biomedical applications.

Our contributions are the following:
\begin{enumerate}
	\item We propose four abilities essential for evaluating retrieval-augmented large language models in the biomedical domain and introduce a benchmark called BIORAB to assess these capabilities. To our knowledge, this is the first benchmark tailored specifically to evaluate these four abilities for RALs in the biomedical domain.
\item we propose a Detect-and-Correct method and a contrastive
learning approach to enhance the RAL’s robustness (unlabeled and counterfactual)  and  negative
awareness;
	\item We evaluated the LLM using the retrieval-augmented method and identified limitations in four key abilities;
	\item We evaluate the RAL on 5 different biomedical tasks over 11 datasets by using  5 LLMs with 3 retrievers.
\end{enumerate}
\section{Results}
\subsection{Results of  RALs  and  backbone LLMs}

We first benchmark various LLMs and RALs on 11 datasets, the results are shown in Table~\ref{con:basic_results_bone}.  In the triple extraction task, we observed that RALs outperformed LLMs (specifically RALs without a retriever) on LLaMA2-13B,MedLLaMA-13B,  LLaMA3-8B and  phi4 14B achieving better performance. For example,  RALs (MedLLaMA 13B with Contriever) enhanced the original MedLLaMA 13B  by  22.37\%, in terms of F1 score on the ADE dataset.  Another finding is that for Qwen2.5-32B, using different retrievers does not lead to significant performance improvement.

RALs have also been evaluated as effective in improving the performance of LLMs across tasks such as link prediction, text classification, and natural language inference.  RALs (LLaMA2 13B with Contriever) enhanced the original LLaMA2 13B  by  0.40\%, in terms of F1 score on the PHarmKG dataset, RALs (MedLLaMA 13B with BM25) enhanced the original MedLLaMA 13B  by  11.86\%, in terms of F1 score on the Hetionet dataset,  RALs (LLaMA2 13B with MedCPT) enhanced the original LLaMA2 13B  by  0.40\%, in terms of F1 score on the Ade-corpus-v2 dataset, RALs (LLaMA2 13B with Contriever) enhanced the original LLaMA2 13B  by  1.67\%, in terms of F1 score on the SemClass dataset, RALs (LLaMA2 13B with MedCPT) enhanced the original LLaMA2 13B  by  6.59\%, in terms of Macro-avg F1 on the BioNLI dataset.

On MedMCQA, our findings differ from other works~\citep{xiong2024benchmarking} as we observed that LLMs outperform RALs in achieving the best performance, we speculate that the reason for this discrepancy lies in the nature of label-not-sensitive tasks, where RALs have the capability to retrieve large corpora such as PubMed~\citep{sayers2021database} or other relevant datasets. In our study, however, our corpus is derived solely from the training set, which may limit the breadth of knowledge accessible to the RALs.

\begin{table*}[ht]
	\centering
	\renewcommand\arraystretch{1.3}
\resizebox{1\textwidth}{!}{%
	\begin{tabular} {c|c|ccc|ccc|ccc|c|c}
		\toprule 
  
	\multicolumn{1}{c}  {}&\multicolumn{1}{c}  {}&\multicolumn{3}{c}  {Triple Extraction}& \multicolumn{3}{c}  {Link Prediction }& \multicolumn{3}{c}  {Classification }&\multicolumn{1}{c}  {QA} &\multicolumn{1}{c}{NL Inference} \\
		%\hline
   LLM &Approach & ADE &  ChemProt & GIT&  PHarmKG &  Hetionet & DS &Ade-corpus-v2 &  SemClass &  SDoH classification &MedMCQA & BioNLI  \\ 
   \hline
     \multirow{4}*{LLaMA2-13B}&BM25~\citep{robertson1976relevance} &  30.93 &   49.11   & 57.61& 97.60    & 82.37   & 75.86  & 95.40  &  75.50 & 63.71 & 40.42  &45.10    \\
     &   Contriever~\citep{izacard2021unsupervised} &  36.06 &  85.40    &  73.55&   \underline{98.00 }  &  77.00  &76.50   &  96.60 &   \underline{79.33 }& 61.18 & 35.52  & 35.12   \\
     &MedCPT~\citep{jin2023medcpt}  & 30.81  &  85.82    & 74.52 &  97.40   &  81.60  & 78.01  &  \underline{ 96.80 }& 78.33  &68.77  & 36.80  &   69.21 \\
     &No Retriever & 34.86  &  77.48    & 58.99 &  97.60   &  80.80  & 77.15  & 96.40  & 77.66  & 72.57 &  41.52 &  62.62   \\
     \hline
     \multirow{4}*{MedLLaMA-13B}&BM25~\citep{robertson1976relevance} & 33.77  &  52.02    & 57.89 &  95.00   &   \underline{ 90.04} &  71.33 &  95.60 & 72.67  & 65.40 & 37.86  & 48.81   \\
     &   Contriever~\citep{izacard2021unsupervised} &  34.58 & 85.69     &  65.88&  97.00   & 77.20   & 75.64  &  95.60 & 77.66  &56.96  &  29.77 & 53.07   \\
     &MedCPT~\citep{jin2023medcpt}  &  31.41 &    80.70  &  \underline{75.24} &   97.40  & 84.40   &  72.84 & 95.40   & 76.16  &66.66  &   33.88&   53.68 \\
     &No Retriever & 12.21 &   50.52   &  42.05&  97.20   &  78.54  &  \underline{80.38}  & 95.40  &  64.00 & 67.71 &  46.47 &   61.07  \\
      \hline
      \multirow{4}*{LLaMA3-8B}&BM25~\citep{robertson1976relevance} & 27.79  &  70.10    & 62.65 & 96.80    & 81.80   & 75.43  & 94.80  & 75.50  & 71.30 & 37.79  & 19.17   \\
     &   Contriever~\citep{izacard2021unsupervised} & 32.70  & 86.91     &  70.96&   96.60  &  73.40  & 78.01  & 94.60  & 75.83  &  72.57&28.11   &   63.85 \\
     &MedCPT~\citep{jin2023medcpt}  &  29.19 &  84.81    & 57.59 &  97.00   & 83.00   & 76.29  & 95.40  &  74.67 & 71.72 & 31.56  &  56.89  \\
     &No Retriever &  7.05 & 21.32     & 74.27 &   97.20  &  81.80  & 80.38  &  93.80 &  73.16 & 57.00 &  55.91 &  6.71   \\
         \hline
      \multirow{4}*{ Phi4 14B} &BM25~\citep{robertson1976relevance} & 36.23  & 81.26     & 60.61 & 97.20    &   88.80 & 60.75  & 88.20  & 74.67  &78.90  & 35.49  & 67.65   \\
     &   Contriever~\citep{izacard2021unsupervised} & 27.66  & 69.70     & 74.84 &  31.60   &  74.20  &53.58   & 90.80  &  75.17 & \underline{81.43}  &33.89   & 56.65   \\
     &MedCPT~\citep{jin2023medcpt}  & 26.35  & 79.48     & 72.55 & 96.80    &   82.80 & 57.38  & 90.60  &  75.00 &  80.16&  27.72 &  85.45  \\
     &No Retriever &  16.56 &   77.23   & 30.09 &   97.60  &   82.20 & 54.05  & 89.20  &  72.50 &74.68  &  \underline{66.18 } &  82.50   \\
         \hline
      \multirow{4}*{ Qwen2.5 32B }&BM25~\citep{robertson1976relevance} & 40.56 & 86.24     & 65.51 &  97.20   &  82.00  &  72.44 & 94.20  & 78.50  & 78.05 & 28.39 &  81.29  \\
     &   Contriever~\citep{izacard2021unsupervised} &  \underline{45.65}  &  87.16    & 73.12 & 97.60    & 75.80   & 77.37  &  94.20 & 76.33  &80.16  &  28.10 &   89.99 \\
     &MedCPT~\citep{jin2023medcpt}  & 42.62  &   87.72   & 69.46 &  96.20   &  82.80  & 71.78  & 95.00  & 76.83  & 72.99 &   28.22&  \underline{91.19 }  \\
     &No Retriever &   26.76&   \underline{ 88.09}   & 68.50 &  90.40   &  81.00  & 76.93  &  94.80 &  48.00 &74.26  & 20.91  &    63.20 \\
           \bottomrule
	\end{tabular}
 }
\vspace{+2mm}
\caption{ Micro F1 results for  triple extraction, link prediction, and text classification, along with Macro-average F1 for question answering and BioNLI, across various approaches on 11 datasets.Underline  indicates the best performance on each
dataset. For complete results, please refer to the Supplementary Materials.} 
\label{con:basic_results_bone}
 
\end{table*}

\subsection{Results  of Testbed 1, 2 and 3}
\label{con:testbed1}

\begin{table*}[ht]
	\centering
	\renewcommand\arraystretch{1.3}
\resizebox{1\textwidth}{!}{%
	\begin{tabular} {c|cc|cc|ccc|c}
		\toprule 
  
\multicolumn{1}{c}  {}&\multicolumn{2}{c}  {Triple Extraction}& \multicolumn{2}{c}  {Link Prediction }& \multicolumn{3}{c}  {Classification}  & \multicolumn{1}{c}{NL Inference} \\
		%\hline
  Corpus & ADE &    GIT&  PHarmKG &  Hetionet   &Ade-corpus-v2 &  SemClass &  SDoH classification &   BioNLI  \\ 
   \hline
   Unlabeled corpus& 14.39       & 1.01 & 97.20  &   78.60    & 93.00 & 6.83  & 72.99  &   61.38   \\
\hline
 Counterfactual corpus (20\%)&   45.06     & 73.03  & 97.40   &  \underline{94.80}  & 95.80  &73.33 & 75.50 & 87.62    \\
      Counterfactual corpus (80\%)&  30.07      & 74.84  & 97.80   &  85.26     &95.00   &75.66    &13.34  &  60.09   \\
      Counterfactual corpus (100\%)&  39.98      &  74.81 & 97.60   &  76.60     &96.80   &77.66    & 0  &   58.52    \\
      \hline
Diverse corpus&   10.08     &  69.80 &97.20  &  75.41      & 96.20&  75.33 &   \underline{ 83.90} & 80.32     \\
\hline
        Labeled corpus &  \underline{45.65 }      &   \underline{75.24}&  \underline{98.00}    &90.40         &  \underline{96.80  }   &  \underline{ 79.33 }& 81.43  &  \underline{91.19}  \\
       None &     26.76     & 42.05 & 97.60  & 78.54    & 96.40   &  77.66  & 74.68   & 63.20 \\

           \bottomrule
	\end{tabular}
 }
\vspace{+2mm}
\caption{Micro F1 results for triple extraction, link prediction, text classification, along with Macro-average F1 for BioNLI, across various approaches on 8 datasets under  \textbf{Testbed 1, Testbed 2, and Testbed 3}. Underline indicates the best performance on each
dataset. Note: we utilize the best RAL in the table~\ref{con:basic_results_bone} as the backbone.}
\label{con:testbed123_results}
 
\end{table*}
We evaluate the model performance based on the  unlabeled corpus, and the results are shown in
Table~\ref{con:testbed123_results}. We have the following observations:

(1) \textbf{RAL utilizing the unlabeled corpus exhibits lower performance compared to RAL utilizing the labeled corpus}. RALs have demonstrated a strong dependence on the labeled corpus, especially on the label-intensive tasks.
For instance, with labeled corpus, the performance of RAL surpasses that of RAL without labeled corpus by 26.41\% on ADE.

(2) \textbf{Even without an unlabeled corpus, RAL still contributes to improving LLM performance in certain tasks}
As shown in Table~\ref{con:testbed123_results}, On Chemprot and Hetionet, RAL utilizing an unlabeled corpus could enhance the original LLM's performance by 30.16\% and 0.06\%, respectively. We speculate that LLMs may possess sufficient knowledge to contribute to enhancing model performance on specific datasets.

We evaluate the model performance based on different counterfactual rates, and the results are shown in
Table~\ref{con:testbed123_results}. We have the following observations:

(1) \textbf{Counterfactual corpus posses a challenge for RALs}. On ADE, counterfactual instances significantly influence the model performance. For instance, when the counterfactual rate is set to 80\%, the triple F1 drops to  around 10\%, showcasing a considerable disparity compared to the triple F1 performance on the labeled corpus.
Similar observations are noted in GIT, PharmKG, ADE-corpus-v2, SemClass, and BioNLI. This suggests that RALs can be easily misled by counterfactual corpus.

(2) \textbf{A lower counterfactual rate may have a reduced impact on RALs}. On Hetionet, we observed that when the counterfactual corpus is set to 20\%, the model performance is better than the factual corpus. We speculate that retrievers have a greater chance of obtaining useful information when the counterfactual rate is lower.

(3)  \textbf{The counterfactual corpus can still contribute to improving LLM performance}. On ADE, GIT, PHarmKG, Hetionet, Ade-corpus-v2, SemClass, BIoNLI. The interesting finding is that even with a  counterfactual corpus, the RAL performance often surpasses the original LLM.  We speculate that the counterfactual corpus may have a beneficial effect on LLMs. Despite the content of the instances being counterfactual, the provided templates still aid in generation.

(4) \textbf{Counterfactual rates and model performance are not inversely proportional.} This finding contradicts human intuition. In some datasets, such as SemClass, when the counterfactual rate is higher, the model performance also improves. This suggests that RALs possess a certain ability to handle counterfactual facts.

% \subsection{Results and Discussion on Testbed3: Diverse Robustness}

% \begin{table*}[ht]
% 	\centering
% 	\renewcommand\arraystretch{1.3}
% \resizebox{1\textwidth}{!}{%
% 	\begin{tabular} {c|cc|cc|ccc|c}
% 		\toprule 
  
% \multicolumn{1}{c}  {}&\multicolumn{2}{c}  {Triple Extraction}& \multicolumn{2}{c}  {Link Prediction }& \multicolumn{3}{c}  {Classification }& \multicolumn{1}{c}{NL Inference} \\
% 		%\hline
%   Corpus & ADE &    GIT&  PHarmKG &  Hetionet   &Ade-corpus-v2 &  SemClass &  SDoH classification &  BioNLI  \\ 
%    \hline
      
%      Diverse corpus  &   39.62     & 69.80  &  97.20  & 75.41     & 96.20  & 75.33     &  83.90  &   \\
%         Labeled corpus & 45.65       &  75.24& 98.00    &90.40         & 96.80     &  79.33 & 81.43   &   91.19  \\
%        None &     26.76     & 42.05 & 97.60  & 78.54    & 96.40   &  77.66       & 74.68   &    63.20 \\
 
%            \bottomrule
% 	\end{tabular}
%  }
% \vspace{+2mm}
% \caption{\added{Micro F1 results for  triple extraction, link prediction, text classification,  and question answering, along with Macro-average F1 for BioNLI, across various approaches on 11 datasets.Underline with green shade indicates the best performance on each
% dataset. For complete results, please refer to the Appendix}~\ref{con:results_of_llms}}
% \label{con:basic_results}
 
% \end{table*}

We evaluate the model performance of diversity robustness, and the results are shown in
Table~\ref{con:testbed123_results}. We have the following observations:

\textbf{The diversity labeled corpus poses a challenge to improve RALs}. We found that RALs consider the knowledge in the diverse corpus as noise, which could potentially impact RAL performance, particularly evident in ADE  datasets.
However, on BioNLI, the diversity labeled corpus could contribute to enhancing the model performance. We speculate that one reason is the retriever we used couldn't retrieve useful information, while another reason could be that the corpus lacks the necessary information.
% \subsubsection{Error Analysis}

% On ADE, we discovered that the Diversity-labeled corpus also leads to redundancy in RAL generation, for instance, in sentence \textit{easily reversible hypoxemia and hypotension induced by nimodipine.}, the expected tail entity is \textit{hypotension}, while RAL regarded the \textit{hypoxemia and hypotension induced by nimodipine.} as the entity.  
% It also struggles with extracting complex entities. For example, in the sentence \textit{clinical, spectroscopic, and imaging abnormalities resolved with discontinuation of metronidazole}, \textit{clinical, spectroscopic, and imaging abnormalities} is considered the ground truth, while RAL regards the entire sentence \textit{clinical, spectroscopic, and imaging abnormalities resolved with discontinuation of metronidazole} as a single entity. In summary, we find that the primary challenge lies in entity recognition, especially in the recognition of tail entities.
% %%%%%%
% On MedMCQA, we observed that error generation primarily stemmed from misjudgment. For instance, in sentence \textit{Question: All of the following muscles are elevators of the mandible EXCEPT:
%  Options: (A) Digastric; (B) Masseter; (C) Medial pterygoid; (D) Temporalis}, the ground truth is \textit{A}, while RAL generates the \textit{D}.  
 In addition to the public dataset, we also used our labeled private dataset, SDoH Classification, which is derived from real-world data. We observed the same conclusion as with the public dataset: the unlabeled dataset and counterfactual corpus influence the performance of RALs. We did not include the our labeled private dataset DS, as RALs without retrievers achieved better performance.

\subsection{Results of Testbed4: Negative Awareness}

We evaluate the model performance of negative awareness, and the results are shown in
Table~\ref{con:test_bed_4}. We have the following observations:

\textbf{RAL poses a challenge to the Negative Awareness}. The true negative awareness rate on PharmKG and BioNLI was zero, and it was only 1.07\% on ADE. Interestingly, the overall performance of fake negative awareness is better than that of true negative awareness. This suggests that RALs still struggle with self-awareness regarding which examples could provide useful information for generations.

\begin{table*}[ht]
	\centering
	\renewcommand\arraystretch{1.3}
\resizebox{0.9\textwidth}{!}{%
	\begin{tabular} {l|l|c|c}
		\toprule 
  
		  Task&Dataset & True negative awareness rate &Fake negative awareness rate\\ 
    \hline
            \multirow{2}*{triple extraction}&ADE&   5.75 & 94.24   \\
	                  % & ChemProt&  19.24 & 77.49\\
                       &GIT&  27.73 & 69.75 \\
             \hline
             \multirow{2}*{link prediction}&PHarmKG&  0.00 & 63.11  \\
	           & Hetionet& 1.71  & 31.33\\
            \hline
            \multirow{3}*{text classification}&Ade-corpus-v2& 68.75    &  70.45  \\
	     & SemClass&1.49   &  99.35  \\
            &  SDoH classification  &   11.18   &  86.22   \\
             % & SDoH classification add aware & 53.11   & 46.00  \\
         % \hline
             % \multirow{1}*{question answering} & MedMCQA&   0.26  &  3.92  \\
          % \cmidrule(rr){2-4}
             \hline
           \multirow{1}*{natural language inference} &  BioNLI&   0.00 &0.38   \\
           \bottomrule

	\end{tabular}
 }
\vspace{+2mm}
\caption{RAL Performance of ADE,  GIT, PHarmKG, Hetionet, Ade-corpus-V2, SemClass, SDoH and BIoNLI on \textbf{Testbed4: negative awareness}.}
\label{con:test_bed_4}
 
\end{table*}

\subsection{Results of our method on unlabeled database}
 
\begin{table*}[ht]
      \Huge
	\centering
	\renewcommand\arraystretch{1.3}
\resizebox{0.6\textwidth}{!}{%
	\begin{tabular} {c|ccc|ccc}
		\toprule 
        
		\multicolumn{1}{c}  {}&\multicolumn{3}{c}  {Ade-corpus-v2}&\multicolumn{3}{c}  {SemClass}\\
		%\hline
		Corpus &  Precision &  Recall & F1& Precision &  Recall & F1 \\ 
   \hline
       Unlabeled corpus&93.00 &93.00&93.00&6.83 &6.83 &6.83   \\
             Labeled corpus& 96.80 &  96.80&  96.80  & 79.33 & 79.33 & 79.33 \\
               None&96.40&96.40&96.40&77.66& 77.66&77.66 \\    
               Detect-and-Correct  & 94.00 & 94.00 & 94.00 & 75.17 & 75.17 & 75.17 \\
           \bottomrule
	\end{tabular}
 }
\vspace{+2mm}
\caption{Results of our Detect-and-Correct method on the text classification task demonstrate improved robustness when using an unlabeled corpus in the retrieval progress. Note: we use the same RAL for different corpus.}
\label{con:unlabed_solved}
\end{table*}

Table~\ref{con:unlabed_solved} presents the results of our Detect-and-Correct method compared to the RAL with different retrieval corpus settings for text classification on Ade-corpus-v2 and SemClass. Our method achieves an F1-score of 94.00\% on Ade-corpus-v2 and 75.17\%  on SemClass, showing competitive performance. The RAL with labeled corpus yields the highest scores, with 96.80\%  and 79.33\%  F1-scores, respectively, while the RAL with  unlabeled corpus performs poorly on SemClass (6.83\%  F1-score). Notably, our Detect-and-Correct method outperforms the  the RAL with an unlabelded corpus, demonstrating its effectiveness in improving robustness. These results highlight the impact of correction-based approaches in text classification tasks. Note: To ensure fairness, we use the best-performing RAL from Table 3 as the backbone for each task.

\subsection{Results of our method on counterfactual database}

\begin{table*}[ht]
      \Huge
	\centering
	\renewcommand\arraystretch{1.3}
\resizebox{0.5\textwidth}{!}{%
	\begin{tabular} {c|c|c}
		\toprule 
            \multicolumn{1}{c} {}&\multicolumn{1}{c}  {PharmKG}&\multicolumn{1}{r}  {BioNLI}\\
		%\hline
		Corpus & F1 & F1 \\ 
   \hline
       % Unlabeled corpus & 41.41 & 41.33  & 41.30 &   \\
       Counterfactual corpus (100\%)&  97.60 &  58.52 \\
       Unlabeled corpus & 78.60 &   61.38 \\
             Labeled corpus & \underline{98.00 }  &  \underline{91.19}      \\
               None & 97.60 &  63.20  \\
               Detect-and-Correct  & 97.80  & 72.77    \\
           \bottomrule
	\end{tabular}
 }
\vspace{+2mm}
\caption{Results of our Detect-and-Correct method on the PharmKG and BioNLI.}
\label{con:ct_solved}
\end{table*}
% Note by Han: Retriever, LLM, and GPT-4 version (for Dectect-and Correct only) combination: BM25, phi4-14B, GPT4o-mini
Table~\ref{con:ct_solved} reports the F1 scores of   retrieval-augmented large language model under five corpus settings across two biomedical benchmarks: PharmKG and BioNLI.  Our proposed Detect-and-Correct framework enhances retrieval by leveraging GPT4 to revise retrieved counterfactual samples prior to model inference. This method substantially improves performance on BioNLI, achieving an F1 score of 72.77\%, outperforming all other retrieval configurations. On PharmKG, although the RAL with  labeled corpus yields the highest F1 score (98.00\%), our method performs comparably (97.80\%), demonstrating the efficacy of counterfactual correction in mitigating retrieval noise.

\subsection{Results of our method on Awareness }
By calculating the True Negative Awareness Rate and Fake Negative Awareness Rate of the best RALs on the SDoH classification, we found that our method, which pre-trains the model using RAL to improve its ability to distinguish between positive and negative instances, achieves a 49.56\% awareness rate, compared to 48.70\% for RAL alone. These results indicate that our approach enhances the model’s ability to differentiate between classes. The improvement suggests that incorporating distinguishing mechanisms strengthens classification robustness.

\subsection{Sampling bias}
 To investigate how sampling bias impacts evaluation, we conducted simulation studies on a two-label classification task. We constructed datasets with different positive-to-negative ratios of 1:1 and 1:5 and reported the results for weighted precision, recall, F1-score, AUROC, and AUPRC. In these experiments, we used the best-performing RAL LLaMA2-13B + MedCPT for Ade-corpus-v2 and LLaMA2-13B + Contriever for SemClass, respectively.
%%%%%%%
Table~\ref{tab:metrics} presents performance metrics across different datasets (Ade-corpus-v2 and SemClass) with varying positive-negative ratios (1:1 vs. 1:5). The reported metrics include Micro Precision, Micro Recall, Micro F1-score, Weighted Precision, Weighted Recall, Weighted F1-score, AUROC, and AUPRC.

 For Ade-corpus-v2, the model performs consistently across both balanced (1:1) and imbalanced (1:5) datasets, with minimal variation in F1-score and AUROC.
For SemClass, there is a significant drop in performance when shifting from a balanced 1:1 ratio (F1-score: 66.50\%, AUROC: 67.40\%) to an imbalanced 1:5 ratio (F1-score: 48.50\%, AUROC: 51.75\%). This suggests that class imbalance significantly affects performance in SemClass, leading to reduced classification robustness.
Sampling Bias and Model Performance:

 The degradation in SemClass performance under imbalance indicates that the model struggles to generalize when one class is underrepresented in training data.
While AUROC remains high in Ade-corpus-v2, the SemClass model under 1:5 imbalance sees a drastic drop in AUROC (67.40\% → 51.75\%), highlighting the difficulty in distinguishing between positive and negative samples.
A similar trend is observed in Weighted F1-score, which drops from 66.08 (1:1) to 35.04 (1:5), showing that over-representing the negative class leads to biased predictions.

\begin{table*}[ht]
\centering
\small
\renewcommand\arraystretch{1.3}
\resizebox{1\textwidth}{!}{
\begin{tabular}{c|c|ccc|ccc|cc}
\hline
\textbf{Dataset} & \textbf{Positive Negative Ratio} & \textbf{Micro Precision} & \textbf{Micro Recall} & \textbf{Micro F1} & \textbf{Weighted Precision} & \textbf{Weighted Recall} & \textbf{Weighted F1} & \textbf{AUROC} & \textbf{AUPRC} \\ \hline
Ade-corpus-v2 & 1:1 & 93.40 & 93.40 & 93.40 & 93.39 & 93.40 & 93.52 & 93.91 & 90.23 \\ 
Ade-corpus-v2 & 1:5 & 93.60 & 93.60 & 93.60 & 93.66 & 93.60 & 93.63 & 92.58 & 90.38 \\ \hline
SemClass & 1:1 & 66.50 & 66.50 & 66.50 & 68.86 & 66.50 & 66.08 & 67.40 & 77.47 \\ 
SemClass & 1:5 & 48.50 & 48.50 & 48.50 & 63.38 & 48.50 & 35.04 & 51.75 & 66.62 \\ \hline
\end{tabular}}
\caption{Performance metrics for different datasets and positive-negative ratios.}
\label{tab:metrics}

\end{table*}

\subsection{Error Analysis}

\subsubsection{Testbed 1}
In this part, we use ADE, GIT, and BioNLI as examples for error analysis. To better understand the impact of the unlabeled corpus on model generation, this part primarily analyzes the  RAL performance on ADE, GIT, and BioNLI, which exhibited the poorest performance among the nine datasets used.  We primarily summarize two error types as shown in Table~\ref{con:error_unlabeled}. We observed that with the unlabeled corpus, RAL tends to generate redundant information and struggles to accurately predict the output, such as the head entity or relation type in the triple extraction task.
% (1) Generate redundant information

\begin{table}[ht]
	\centering
	\renewcommand\arraystretch{1.3}
	\scalebox{0.8}{
	\begin{tabular} {c|cccc}
		\hline 
		Error type&Dataset& Input sentence  & Expected output& Error output\\ 
		\hline	
       \multirow{5}*{\makecell[l]{Redundant \\ information}}&ADE & \makecell[l]{the fourth patient showed rls  symptoms \\that were initially   caused by a 20-mg\\ daily  olanzapine dosage   and were later\\ mitigated when olanzapine was \\ reduced and ropinirole was administered.}  & \makecell[l]{\{olanzapine, \\ 
       dosage, \\
       20-mg daily\}\\ }  &  \makecell[l]{\{olanzapine, \\ 
       dosage, \\
       rls symptoms that \\ were initially caused by \\ a 20-mg dail\}\\ }  \\
         \cmidrule(rr){2-5}
        &GIT & \makecell[l]{inactivation kinetics of vacterial \\ glycerol dehydratase  (ec 4.2.1.30) in\\ the course of its reaction with  \\adenosylcobalamin  (adocbl) and\\ its analogs were investigated..}   &  \makecell[l]{  \\ 
       glycerol dehydratase}  &  \makecell[l]{ \\ 
       adenosylcobalamin.. \\retrieved sentence: \\ glycerol dehydratase}  \\
         \cmidrule(rr){2-5}
         & BIONLI&  -- &  negative &  negative retrieved sentence..\\
         \hline
    \multirow{3}*{\makecell[l]{Error \\ generation}}&ADE &  \makecell[l]{four patients receiving high-dose \\ tamoxifen for greater than  \\ 1 year have demonstrated \\ similar retinal changes.}   &\makecell[l] {(tamoxifen, \\dosage, \\ high-dose)} & \makecell[l] {(tamoxifen, \\effect, \\ retinal changes.)}\\
  \cmidrule(rr){2-5}
    &GIT &  \makecell[l]{inactivation of serum  alkaline \\ phosphatase  by  adrenaline  \\ and related substances}.  & \makecell[l] {(adrenaline, \\inhibits, \\ alkaline phosphatase)} &  \makecell[l] {(alkaline phosphatase, \\interacts with, \\ adrenaline)} \\
      \cmidrule(rr){2-5}
&BIONLI &  --- & positive &negative  \\
         \hline
       
	\end{tabular}
 }
		\caption{Error cases of Unlabeled Robustness. In BioNLI, we have not included the input sentence in this table due to the excessive length of the sentences.}
	\label{con:error_unlabeled}
\end{table}

\subsubsection{Testbed2}
In this part, we use PharmKG and SemClass as examples for error analysis. For the query "What is the relationship between ethanol and GABRA1?" in PharmKG, the ground truth label is "Interactions". However, when the retrieved corpus contains 80\% noise, the generated answer is  "Chemical-Gene", indicating that excessive noise in retrieval can mislead the model and result in incorrect predictions.
For example in the SemClass,  the model is asked to determine whether the triple (nitrous acid, INHIBITS, protein) is correctly inferred from the sentence  "Inactivation of protein in poliovirus by nitrous acid." The ground truth label is True, indicating that the relationship exists. However, the model predicts False, suggesting it fails to recognize the inhibitory relationship expressed in the sentence. This misclassification may stem from the model struggling with implicit linguistic cues, misunderstanding passive voice structures, or failing to generalize inhibition-related terms effectively.

\subsubsection{Testbed3}

\label{con:testbed1}
In this part, we use ADE and MedMCQA as examples for error analysis. On ADE, we discovered that the Diversity-labeled corpus also leads to redundancy in RAL generation, for instance, in sentence "easily reversible hypoxemia and hypotension induced by nimodipine.", the expected tail entity is "hypotension", while RAL regarded the "hypoxemia and hypotension induced by nimodipine." as the entity.  
It also struggles with extracting complex entities. For example, in the sentence "clinical, spectroscopic, and imaging abnormalities resolved with discontinuation of metronidazole", "clinical, spectroscopic, and imaging abnormalities" is considered the ground truth, while RAL regards the entire sentence "clinical, spectroscopic, and imaging abnormalities resolved with discontinuation of metronidazole" as a single entity. In summary, we find that the primary challenge lies in entity recognition, especially in the recognition of tail entities.
On MedMCQA, we observed that error generation primarily stemmed from misjudgment. For instance, in sentence "Question: All of the following muscles are elevators of the mandible EXCEPT:
 Options: (A) Digastric; (B) Masseter; (C) Medial pterygoid; (D) Temporalis", the ground truth is "A", while RAL generates the "D".

\section{Discussion}
\subsection{Testbeds and our methods}
The results demonstrate that the reliance on labeled versus unlabeled corpora significantly impacts model performance, particularly in label-intensive tasks like ADE, where labeled data improves results by 26.41\%. Despite this, RAL still enhances LLM performance even when relying on an unlabeled corpus, as seen in Chemprot and Hetionet, suggesting that LLMs contain useful intrinsic knowledge. Counterfactual corpora introduce challenges, with high counterfactual rates (80\%) reducing F1 scores to around 10\%, indicating model susceptibility to misleading information.  Interestingly, counterfactual corpora sometimes enhance performance, as observed in SemClass, suggesting that even incorrect instances can contribute useful templates for generation. This phenomenon implies that RALs can handle counterfactual information to some extent, contradicting initial expectations. Additionally, diversity within labeled corpora poses a challenge for RALs, particularly in ADE, where the model struggles to distinguish noise from valuable knowledge. Negative awareness remains a weakness, with RAL performing poorly in distinguishing false negatives, as evident in PharmKG and BioNLI. However, our Detect-and-Correct method significantly improves text classification robustness, outperforming the unlabeled corpus in SemClass. These findings highlight the importance of correction-based approaches and tailored retrieval mechanisms for optimizing model performance across diverse datasets.Otherwise, by leveraging contrastive learning methods, RAL can enhance its ability to distinguish between positive and negative instances, thereby improving its awareness and overall performance.

\subsection{Future work}
   To improve the Unlabeled Robustness, we plan to  explore ways to improve label generation through pretraining, fine-tuning, or retrieval-based methods to enhance the model’s labeling capability. Another promising direction is training retrieval models to quickly adapt to new label-intensive tasks with minimal supervision, thereby improving generalization in low-resource settings. These efforts will further strengthen the robustness of RAL when handling unlabeled data, making it more effective for real-world biomedical applications.

 To enhance RALs' ability to handle counterfactual corpora, we plan to explore the following directions. First, we aim to develop retrieval strategies that incorporate uncertainty estimation, allowing RALs to weigh retrieved instances based on their likelihood of being counterfactual. Second, we propose using contrastive objectives to train RALs to distinguish between factual and counterfactual instances, thereby improving retrieval robustness against misleading examples.

  To enhance retrieval robustness, incorporating uncertainty-aware filtering and hybrid retrieval (dense + sparse methods) could help reduce reliance on misleading information. Addressing negative awareness requires contrastive learning for retrieval filtering and adversarial training on counterfactual data to improve the model’s ability to recognize false negatives. Additionally, domain-specific adaptation can be improved by pretraining on biomedical corpora and fine-tuning with structured knowledge graphs to enhance factual consistency. Future work should explore these strategies to make RALs more reliable in complex biomedical tasks. 

 To quantify hallucinations, evaluation testbeds should include biomedical fact-verification benchmarks that measure accuracy in factual claims. Metrics such as Retrieval Grounding Score can assess how well model outputs align with retrieved evidence, while Hallucination Rate can track the proportion of responses containing unverifiable or incorrect claims. Implementing uncertainty-aware retrieval filtering would further help reduce the inclusion of low-confidence or misleading retrievals. Future research could also explore multi-hop retrieval reasoning, where RALs retrieve multiple supporting documents to validate complex biomedical claims. These techniques would significantly enhance the reliability of RALs in clinical decision support, biomedical literature analysis, and drug discovery tasks.

 \added{Finally, while our study advances methods for retrieval-augmented learning in biomedical domains, it does not include comparative evaluations with other advanced large language models. Addressing this limitation in future work will be important to assess the broader applicability and generalizability of our findings.}

\section{Materials and Methods}
In this part, we assess RAL's performance across various biomedical NLP tasks, analyze its efficacy on four proposed testbeds, and discuss its abilities.
\subsection{Settings and Dataset}
We evaluated three state-of-the-art LLMs: LLamA2-13B, MedLLamA-13B, LLaMA3 8B,   Phi4 14B and Qwen2.5 32B, along with three retrievers: BM25, Contriver, and MedCPT. We considered five biomedical NLP tasks: triple extraction, link prediction, text classification, question answering, and natural language inference, across nine datasets: ADE, ChemProt, GIT, PHarmKG, Hetionet, Ade-corpus-v2, SemedCLass, MedMCQA, BioNLI, DS  and SDoH. The data statistics are shown in Table~\ref{con:data_stastics}. The experiments were conducted using A100 GPUs.
\subsubsection{Three retrievers}

The three retrievers—BM25, Contriever, and MedCPT—employ a structured process to retrieve the most relevant instances for enhancing LLM predictions. This process consists of three main steps: 
1)The retrievers first compute the vector representation of the given input sentence using their respective retrieval models. BM25 uses term-based weighting, while Contriever and MedCPT rely on dense embeddings generated through neural networks. 2) Each retriever then calculates the vector representation for every instance in the retrieval corpus. BM25 relies on traditional lexical matching, while Contriever and MedCPT employ deep learning-based embeddings to encode semantic information. 3)    Finally, the retrievers measure the similarity between the input sentence vector and each instance vector in the corpus. The most relevant instances, determined by the highest similarity scores, are selected and fed into the LLM alongside the input sentence as examples, helping the model generate more accurate predictions.

\begin{table}[ht]
	\centering
	\renewcommand\arraystretch{1.3}
	\scalebox{0.8}{
	\begin{tabular} {c|cccc}
		\hline 
		&Dataset& train &test& dev\\ 
		\hline	
       \multirow{3}*{Triple extraction}&ADE~\citep{gurulingappa2012ade} &  4,970 & 2,130 & -- \\
        &ChemProt~\citep{taboureau2010chemprot} &  4,001&3,355&2,366\\
         & GIT~\citep{li2023benchmarking}& 3,734&465&492\\
         \hline
        \multirow{3}*{Link Prediction}&PHarmKG~\citep{zheng2021pharmkg} &  4,000 & 500 &  500 \\
         & Hetionet~\citep{himmelstein2017hetionet} & 4,000  &  500&500  \\
          &  SDoH   &  3964   &  464   &--  \\
          \hline
     \multirow{3}*{Text classification} &Ade-corpus-v2~\citep{gurulingappa2012ade} & 4,000  &500&500   \\
         &  SemdClass~\citep{vasilakes2018bionli} &  2,400 & 600 &600  \\
           &   DS   & 2127  &   237 & --\\
          \hline
            \multirow{1}*{Question answering}&   MedMCQA~\citep{pal2022medmcqa} &   34,994 & 4,183 &4,183  \\
             \hline
          \multirow{1}*{Natual language inference}  &   BioNLI~\citep{bastan2022bionli}  &5,544   & 6,308 &12,807  \\
        % MedQA~\citep{jin2021disease}  & 10,178  &1,273&1,272 \\
		\hline
	\end{tabular}
 }
		\caption{Data Statistics for  the datasets we used in this work}
	\label{con:data_stastics}
\end{table}

\subsubsection{Triple Extraction Dataset}  
In this paper, we utilized ADE, Chemprot, and GIT as the foundational datasets.

\begin{enumerate}
 \item ADE  is extended from relation extraction task to triplet extraction task in this paper. All sentences either describe the effect of the drug or the dose of the drug. Thus, the triplets consist of (head entity: drug, relation type: effect, tail entity: effect\_description) and (head entity: drug, relation type: dosage, tail entity: dose\_description). Among all triplets, there are only two relation types: effect and dosage.
  \item ChemProt: The Chemical Protein Interaction Corpus comprises 2432 PubMed abstracts annotated with chemical-protein interactions, encompassing 23 distinct interaction relations. Building upon prior research~\citep{sun2022mrc4bioer}, the corpus exclusively considers sentence-level instances, with a particular focus on five prominent interaction types for classification: CPR3, CPR4, CPR5, CPR6, CPR9.
   \item GIT is a high-quality biomedical triple extraction dataset for non-drug
therapies, characterized by its high-quality annotations and comprehensive coverage of relation types. It includes 22 relation types from SemMedDB.
\end{enumerate}

\subsubsection{Link Prediction}
In this paper, we utilized PHarmKG and  Hetionet as the foundational datasets in the link prediction task. 

\begin{enumerate}
   \item  PHarmKG is a knowledge graph to describe the relationship among genes, drugs, and diseases. In this work, we aim to predict the four mentioned relation types (Interactions, Disease-Gene, Disease-Chemical,  Chemical-Gene) between two entities. During the huge quantity of triples in the PHarmKG,  we randomly select  4,000 samples from the source training set for training, 500 samples from the source testing set for testing, and 500 samples from the source validation set for validation.

   \item  Hetionet is an integrative network of disease, which includes   46 relation types. In our paper, we randomly select  4,000 samples from the source training set for training, 500 samples from the source testing set for testing, and 500 samples from the source validation set for validation.
\item  Dietary supplement (DS).   This task determines whether a dietary supplement is positively or negatively associated with a specific event, or if there is no direct relationship. It facilitates the identification of cause-effect relationships or therapeutic uses of dietary supplements. For example, the task would identify a negative association between “ginseng” and “nausea.

\end{enumerate}

\subsubsection{Text Classification}
In this paper, we utilized Ade-corpus-v2  and SemdClass as the foundational dataset in the text classification task. 
\begin{enumerate}
   \item  Ade-corpus-v2 dataset is designed for classifying whether a sentence is ADE( Adverse Drug Reaction)-related (True) or not (False). In our paper, we randomly select 4,000 instances for training, 500 for testing, and 500 for validation.
   \item  The SemdClass, aims to understand whether the provided triple belongs to the given sentence or not. It includes two classes, False and True.

   \item  Social Determinants of Health (SDoH) focuses on classifying Social and Behavioral Determinants of Health (SDoH) snippets. The dataset consists of 2,364 annotated notes, categorized into twenty broad SDoH categories, covering various social factors that influence health outcomes. These categories include employment status, work conditions, access to food, housing status, income status, health literacy, insurance status, access to transportation, maltreatment history, environmental conditions, housing quality, neighborhood security, crime/incarceration history, social discrimination, social isolation, social support, immigrant status, substance abuse, physical activity, and sleep quality.
\end{enumerate}

\subsubsection{Questing Answering and Natual Language Inference}
In this paper, we utilized MedMCQA as the foundational dataset in the question-answering task and used BioNLI as the dataset of natural language inference.

\begin{enumerate}
  \item  MedMCQA is a multi-choice question-answering dataset that designed to address the medical entrance exam questions.  In this work, we opt for the five-choice version (A, B, C, D, E).

   \item  BioNLI aims to understand whether the provided hypothesis is consistent or adversarial to the premise.
\end{enumerate}

\subsection{BioRAB: Biomedical Retrieval-Augmented Generation Benchmark}
% In this section, we first introduce the working flow about the BioRAG models. Next, we introduce the specific retrieving augmented generation abilities and the building progress of relevant testbeds. Lastly, we present the evaluation metrics.
In this part, we begin by outlining the operational flow of the RALs. Following this, we introduce the proposed four abilities and the building progress of four relevant testbeds. Finally, we introduce the evaluation metrics employed to assess performance.

\begin{figure*}[t]
        \centering
        \includegraphics[width=1\columnwidth]{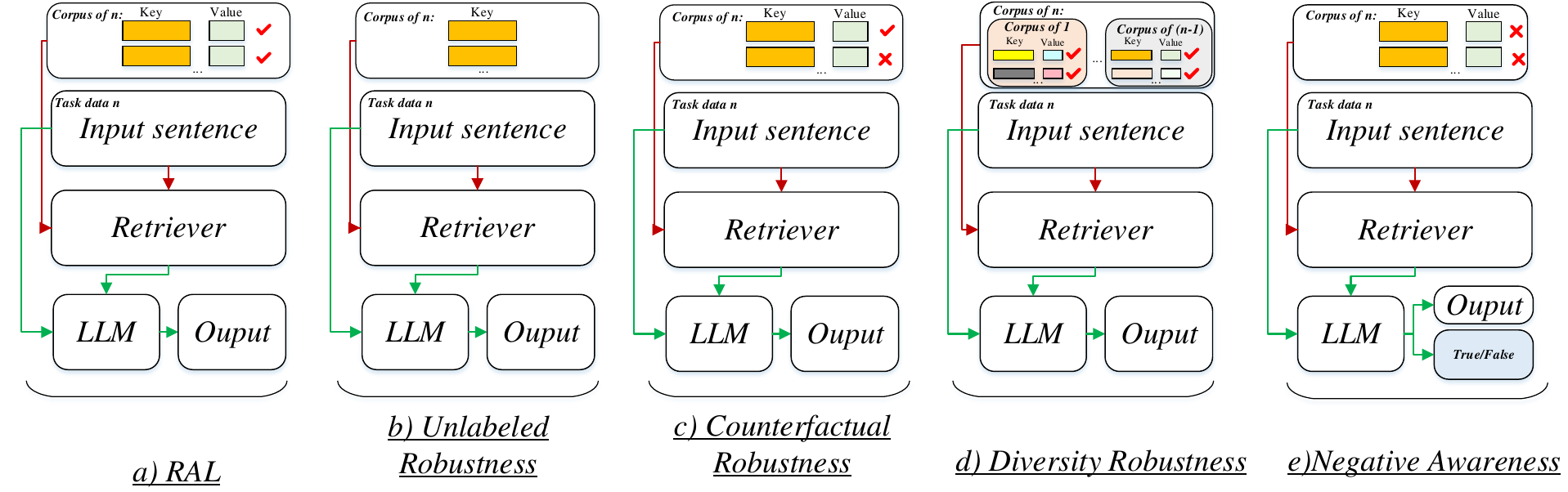}
	\caption{ Overview of four testbeds on \textsc{BIORAB}. $n$ refers to the special dataset for each task, such as ade-corpus-v2 (text classification), and PHharmKG (link prediction). In (d), the corpus of $n$ refers to the set that includes the task datasets but excludes the training set of $n$.
 In (e), to distinguish the difference between "Output" and "True/False", the "Output" is defined as the expected output for different tasks, for example, in the triple extraction task, the output is the triple. "True/False" refers to "the retrieved example is a negative
example or the retrieved example is not a negative example." In our work, the n corpus of n refers to the training set of n. } 
	\label{con:whole_framework}
\end{figure*}

% Given  a biomedical sentence, we aim to generate the triple $(h,r,t)$ from the  

\subsubsection{RAL Working Flow}
To solve the hallucination problem, RAL is proposed to retrieve the external knowledge from the corpus and improve the LLM performance. Generally, as shown in Figure~\ref{con:whole_framework}(a), retrieved corpus needs to be constructed initially, in numerous question-answering RAL models,  the  corpus  primarily originates from the unlabeled open source such as PubMed, Textbook. However, for some label-sensitive tasks, such as triple extraction, the unlabeled open source may be invalid. In our work, the corpus is defined as the training set for the relevant task. For instance, as illustrated in Figure~\ref{con:whole_framework}(a), if "n" denotes the PHarmKG, each key corresponds to a sentence  \textit{what is the relationship between the head entity and tail entity? } in its training set, while the corresponding value denotes the relevant label \textit{relationship} for that key. In  the second step, the retriever is used to obtain the relevant (key, value) pairs from the corpus based on the input sentence. At last, the retrieved (key, value) pairs with the input sentence are fed into the LLM to generate the expected output.
% Note: in our work, the retriever is used in the training, validation, and testing progress,  it is tasked with obtaining the relevant key-value pair for each sentence in the train, valid, and test set. 
For each instance $X$ of each "n", there are three components: Instruction $I$, context $C$, and response $R$. For example, in the training dataset  of ade-corpus-v2 (classification task), if the label of a sentence $S$: \textit{She had been administered tacrolimus for prophylaxis of graft-versus-host reaction} is \textit{False} in the $X$, $I=$\textit{You are an excellent linguist. The task is to predict whether this sentence is True or False. Examples: context: The hemangioma regressed markedly 6 weeks after the procedure and serous retinal detachment showed marked resolution.response: False}, $C=S$, $R=False$.

\subsubsection{Four Abilities of BioRAL}
Despite the  RAL has achieved considerable success in solving the hallucination problem, in the biomedical domain, the ability of RAL is underexplored.
Firstly,  
not all tasks have vast labeled corpora. While many research endeavors employ the training set as the corpus, they still encounter limitations when contrasted with larger corpora.
% not all tasks have labeled corpus that are ready to be retrieved (as shown in Figure~\ref{con:whole_framework}(b)), despite many research work utilize the training set as the corpus, it still have limitation vy comparing the huger corpus.  , as illustrated in Figure~\ref{con:whole_framework}(b)
%%%%%%%%%%
If a RAL can achieve similar performance to the RAL that utilizes labeled corpus, it would demonstrate the former's ability to operate effectively without relying on labeled data.
For another,  the RAL may easily be misled by incorrectly labeled information (as shown in Figure~\ref{con:whole_framework}(c)).   
Furthermore, RALs may possess the capability to obtain useful information from labeled corpora of other tasks (as shown in Figure~\ref{con:whole_framework}(d)).  
However, retrieving knowledge from labeled corpora of other tasks may introduce noise and potentially mislead the generation process.
%%%%%%%%%%%%%
Finally,  when the retriever retrieves mislabeled (or counterfactual) information, the RAL may possess the ability to discern that the retrieved knowledge is not conducive to output generation (as shown in Figure~\ref{con:whole_framework}(e)).   To this end, we built the Biomedical Retrieval-Augmented Generation Benchmark (BIoRAB) to evaluate the ability of RAL in the biomedical domain, and we proposed 4 testbeds to test these abilities. In the next, we will detail these four abilities and how to construct the testbeds.

% \subsubsection{unlabeled Robustness (UR)}
\paragraph{Unlabeled Robustness (UR):} Not all tasks have vast labeled retrieval corpus, therefore, for each task, the retriever must gather information from unlabeled corpora, while the RAL may still have the ability to generate the expected results. 
To evaluate the efficacy of RAL in this regard, we introduce our proposed UR testbed.
Specifically, as shown in Figure~\ref{con:whole_framework}(b), the corpus of "n" is defined as the training set without value(label) for the "n". The retriever retrieves the relevant information from this unlabeled corpus.  After that, the retrieved Key with the input sentence is fed into the LLM.  For example, in the training dataset of  ade-corpus-v2 (classification task), if the label of a sentence $S$: \textit{She had been administered tacrolimus for prophylaxis of graft-versus-host reaction} is \textit{False} In the $X$, $I=$\textit{You are an excellent linguist. The task is to predict whether this sentence is True or False, retrieved sentence: A macrophage activation syndrome, possibly related to methotrexate toxicity, developed in a boy with systemic juvenile rheumatoid arthritis}, $C=S$, $R=False$.

% \subsubsection{Counterfactual Robustness (CR)}

\paragraph{Counterfactual Robustness (CR):}
Constructing a high-quality annotation corpus is challenging work, as it often involves dealing with incorrect data labeling.  In our work, these mislabeled instances are called counterfactual instances.
% In the labeled corpus, noise can manifest as mislabeled instances, while in the unlabeled corpus, noise refers to instances that are related to the input sentence but lack information about the expected output.
% In our work, we mainly focus on the noise in the labeled corpus.
In the condition of the mislabeled corpus, the RAL may have the ability to avoid negative information.
%%%%%%%%%%%%%
To validate the counterfactual robustness, we introduced our CR testbed. Specifically, as shown in Figure~\ref{con:whole_framework}(c), when constructing the corpus of n, we set the negative rate to be $20\%$ or $80\%$ or $100\%$, corresponding to $20\%$ or $80\%$ or $100\%$  of instances being wrongly labeled. An example of incorrect annotation in a classification dataset would be if there are two labels, "True" and "False." If the true class of one instance is "True," then its incorrect annotation would be "False".
Subsequently, the retriever is tasked with retrieving relevant information from this corpus.  The retrieved information, along with the input sentence, is fed into the LLM to generate the output.

\paragraph{Diverse Robustness (DR):}
% \subsubsection{Diverse Robustness (DR)}
Diverse Robustness refers to the ability to incorporate diverse information from various task corpora. On one hand, in numerous scenarios, the corpus from other tasks may contain valuable information to aid in generation. For instance, in the task of triple extraction, if a suitable triple extraction corpus is unavailable, the question-answering corpus may assist in extracting the necessary information. On the other hand, different tasks may introduce noise that could potentially impede the performance of the RAL.
To generate better output, it is necessary for RAL to have the ability to retrieve diverse information.  So, we introduce our  DR testbed, as shown in Figure~\ref{con:whole_framework}(d), 
when constructing the corpus of "n", it incorporates corpora from other tasks. For instance, if "n" refers to the Chemprot (triple extraction task), the corpus of "n" includes corpora from tasks such as GIT (triple extraction task), PHarmKG (link prediction task), and so on. Next, the retriever is required to extract the pertinent information from the diverse corpus. Subsequently, the retrieved information, along with the input sentence, is fed into the LLM to generate the output.

\paragraph{Negative Awareness (NA):}
Negative Awareness evaluates the ability of LLMs to discern whether the retrieved information is negative (it is not conducive to the expected output). In real-world scenarios, if the retriever obtains negative information and the LLM can identify it as such, the LLM can then seek out more useful information to aid in generation based on this feedback. So, we introduce our  NA testbed, as shown in Figure~\ref{con:whole_framework}(e), we designate all values in the corpus of "n" as incorrect labels.  After obtaining the retrieved documents from the corpus, the model is expected to produce two types of output. Firstly, task-based output, such as in the task of triple extraction, the output should be triple. Secondly, the model should also provide a judgment on whether the retrieved knowledge is negative or not.

\subsubsection{Evaluation Metrics}

 \paragraph{Task-based Metrics:}
In the triple Extraction task, same as BiomedRAG~\citep{li2024biomedrag}, triple is regarded as correct when its relation type, the head entity, and the tail entity are all correct. For example, in the sentence: \textit{Infusion of prostacyclin (PGI2) reportedly attenuates renal ischemic injury in the dog and the rat.},  triple \textit{<Infusion, treats, rat>} is regarded as correct while \textit{ <injury, treats, rat>} is not. 
%%%%%%%%%%%
We evaluated all the models and reported the evaluation metric, including Micro Precision, Recall, and F1-score. For the text classification, link prediction, and question answering task, we follow the same evaluation metrics as triple extraction. 
For the natural language inference task, we use the same evaluation metric (Macro F1) as the BioNLI.

\paragraph{Negative Awareness Metrics:}
To assess negative awareness in our study, we define a negative instance as a mislabeled instance. In the first, we need to evaluate the model performance using mislabeled examples. For instance, in the ade-corpus-v2 classification data, with two labels "True" and "False", this evaluation gauges the performance of "True" or "False" predictions.

Typically, in the RAL framework, if the retrieved example contains the input sentence and its expected output, the LLM should achieve 100\% performance when tested with the input sentence. Despite all instances in the retrieval corpus being mislabeled, the LLM may still generate the correct output when utilizing these examples. In our experiments, we also investigate this aspect. 
Building on this discovery, we delineate two types of negative instances:

\begin{enumerate}
  \item \textbf{True negatives}: When the negative instance is provided to the LLM along with the input sentence, resulting in the incorrect output. In this scenario, the number of input sentences is denoted as $l^t$. 
  \item \textbf{False negatives}: When the negative instance is presented to the LLM alongside the input sentence, leading to the correct output. In this case, the number of input sentences is represented as  $l^f$.
\end{enumerate}

%%%%%%%%
At the same time, we also expected the LLM could output \textit{True - The retrieved example is negative example} or \textit{False- The retrieved example is not a negative example}  by providing a specific instruction \textit{Please determine whether the retrieved example constitutes negative information. If it is negative, please output False; if it is not negative, please output True}  for each input sentence.
For an input sentence that has false negative examples, if the LLM could output \textit{False - The retrieved example is not a negative example}, it demonstrates that the LLM recognizes the example as a false negative. After the judgment of LLM, The count of input sentences with "false negative examples" is denoted as  $f$.
%%%%%%%%%
For an input sentence that has true negative examples, if the LLM could output \textit{True - The retrieved example is a negative example}, it demonstrates that the LLM recognizes the example as a true negative. After the judgment of LLM, the count of input sentences with "true negative examples" is denoted as  $t$.
\textbf{So} the true negative awareness rate is calculated by $t/l^t$, and the false negative awareness rate is calculated by $f/l^f$.

\subsection{Improving the Robustness and  Awareness of RALs}
In this part, we mainly introduce our method to improve the Robustness(Counterfactual and unlabeled) and negative awareness ability of the RAL.
% \added{---------}

% \subsubsection{\added{Improving the Robustness of RALS}}
\subsubsection{Improving the Counterfactual Robustness of RALs}
In the context of mitigating mislabeled instances within a retrieved corpus, we propose a two-step approach consisting of a detection phase followed by a correction phase. This methodology ensures that errors introduced during data annotation or retrieval are systematically identified and rectified, thereby improving the quality of the corpus used for downstream tasks. In this work, we primarily focus on developing the testbed and therefore adopt a straightforward approach to address this issue. The investigation of more advanced methods is left for future work.
% \textbf{Mislabeled Instance Detection and Instance Correction:}
More specifically,
The first step in our approach involves developing a mislabeled instance detector that systematically identifies erroneous labels.  Following the detection phase, the mislabeled instance corrector is responsible for modifying the identified erroneous labels.  Specifically, for example,  in the case of a text classification task, we leverage in-context learning (ICL) with large language models (LLMs) (e.g., GPT-4) to assess the correctness of assigned labels, and using the internal knowledge to modify it. The prompt employed for this process is as follows:

\textit{This is a text classification task. Please determine whether the label assigned to the input sentence is correct. If the label is incorrect, please provide the correct label.}

Through this mechanism, the model systematically reviews each labeled instance and flags those suspected to be mislabeled.
By using the insights from the LLM’s evaluation, the incorrect labels are revised accordingly.
\subsubsection{Improving the Unlabeled Robustness of RALs}
In the case of an unlabeled corpus, to enhance robustness in the absence of labeled data, we also leverage  ICL with LLMs to assign labels. This is achieved by providing task-specific instructions. The detailed prompt is as follows (using text classification as an example):

\textit{This is a text classification task aimed at determining whether a given sentence is related to an Adverse Drug Event (ADE). Please assign a label to each provided sentence to support this task}

\subsubsection{Improving the Awareness of RALs}
To enhance the negative awareness of retrieval-augmented language models (RALs), a key direction is to improve the model’s ability to distinguish between positive instances in the corpus—which contribute to generating the correct answer—and negative instances, which do not.

For each input sentences $x$, there are hundreds or thousands of instance in the corpus  not related to  answer gernation. Based on this observation, we design model $S$  to help  distinguish which instance are helpful for the final generation of the input $x$.  This model takes the following information as input: (1) input sentence $x$; (2) positive  instance $C^p=\{e^p\}$,
% \lifu{need a better math notation} 
which is constructed based on the relevance of $x$,  in general, $x$ itself serves as the most informative positive instance.; (3) negative entity set $C^n=\{e^n\}$, which is created by randomly sampling the instances from the corpus is not included in $C^p$.
%%%%%%%%%%%%
To differentiate the positive and negative entities, $S$ utilizes the LLM model $f(.)$, such as llama models to extract a semantic representation $f(x)$, $f(e^p)$, $f(e^n)$ for  $x$, $e^p$, $e^n$, respectively, and select the positive instances by measuring their distance to the input sentences. $S$ is optimized with the following triple loss:
%. The BERT-based entity selection model $S^e$ and semantic representation model $f(.)$ are then optimized by minimizing the triple loss,
\begin{align}
\max(\|f(x)\!\!-\!\!f(e^p)\|\!-\!\|f(x)\!\!-\!\!f(e^n)\|\!+\!\alpha,\!0) \nonumber
\end{align}
where $\|.\|$ denotes the Euclidean Distance and $\alpha$ is a margin parameter, which we set to 1 as default. During training, the triplet loss reduces the distance
between $f(x)$ and $f(e^p)$ while enlarging the
distance between $f(x)$ and $f(e^n)$.  Once the $f(.)$ is trained,  it is subsequently used for instruction tuning in the downstream task. For example, in the task SDoH classification , the trained $f(.)$ based on Phi4 , is utilized to fine-tune the model on the classification dataset.

\section{Acknowledgements}
This work was supported by the National Institutes
of Health’s National Center for Complementary
and Integrative Health grant number R01AT009457, National Institute on Aging grant number
R01AG078154, National Cancer Institute grant number R01CA287413 and Food and Drug Administration grant number U01FD008720. The content is solely the responsibility of the authors and does not represent the official views of the National Institutes of Health. 

\section{Data and Materials Availability}
All data needed to evaluate the conclusions of this paper are present in the paper and/or the Supplemental Materials.The code and data are available at Github: \href{https://github.com/ToneLi/ToneLi-Evaluating-Retrival-LLM-in-Biomedical-Domain?tab=readme-ov-file}{Code and Data in Github}.
Zenodo: \href{https://zenodo.org/records/17398149?token=eyJhbGciOiJIUzUxMiJ9.eyJpZCI6IjJkYTY5NjQzLTVhZjktNDg1OC1hNTJmLTQxMzljM2VhM2Q0YyIsImRhdGEiOnt9LCJyYW5kb20iOiJkODY4MmJjZDg2YjNmMmVmOGNmODI0YWVhOGFlOTY1YSJ9.Qd_p5IuMAWp3B5JoxCM2cwZwmvkk9dcPhj8dZPa3Cyq6Y_eBpKsmgcpj1K_XtbBgYgUZSAgqgXDPBJxjGYrzbQ}{Code and Data in Zenodo}.
\section{Author Contributions}

\begin{enumerate}
  \item Mingchen Li:  Digging for ideas, doing related work, experiments (PharmKG,  Ade-corpus-v2, semclasss, BIoNL), coding,  Methodology, Model Design, Writing – original draft, Paper revision.
  \item Zaifu Zhan:   experiments (ADE, ChemProt, GIT), experiment discussion.
  \item Han Yang:  experiments (MedMCQA), experiment discussion.
    \item Yongkang Xiao:  experiments (Hetionnet), experiment discussion.
     \item Huixue Zhou:  experiments (Ade-corpus-v2, semclasss), experiment discussion.
     \item Jiatan Huang:  experiments (SemClass), code, data, and model organization, experiment discussion.
   \item Rui Zhang: Supervision, idea and method exploration, Writing – original draft, Writing – review \& editing. 
\end{enumerate}

% \begin{itemize}
%   \item Mingchen Li:  Digging for ideas, doing related work, experiments, coding,  Methodology, Model Design, Software, Writing – original draft, Paper revision.
%   \item Halil Kilicoglu:  paper revision,  medical knowledge support.
%   \item Hua Xu:  paper revision, medical knowledge support.
%    \item Rui Zhang: Supervision, idea and method exploration, Writing – original draft, Writing – review \& editing. 
% \end{itemize}

\section{Competing Interests}
The authors declare no competing financial or non-financial interests.

% \section{Supplementary Materials}

% the data and code are available at:
% \url{https://anonymous.4open.science/r/ToneLi-Evaluating-Retrival-LLM-in-Biomedical-Domain-86CD/README.md}

\bibliography{science_template}

\begin{thebibliography}{99}

\bibitem{wu2024pmcllama}
C. Wu, W. Lin, X. Zhang, Y. Zhang, W. Xie, Y. Wang, PMC-LLaMA: Toward building open-source language models for medicine, \textit{Journal of the American Medical Informatics Association (J. Am. Med. Inform. Assoc.)} \textbf{31}(9), 1833--1843 (2024).
\bibitem{robertson1976relevance}
S. E. Robertson, K. Spärck Jones, Relevance weighting of search terms, 
\textit{Journal of the American Society for Information Science (J. Am. Soc. Inf. Sci.)} 
\textbf{27}(3), 129--146 (1976).

\bibitem{singhal2025toward}
K. Singhal, T. Tu, J. Gottweis, R. Sayres, E. Wulczyn, M. Amin, L. Hou, K. Clark, S. R. Pfohl, H. Cole-Lewis, D. Neal, Q. M. Rashid, M. Schaekermann, A. Wang, D. Dash, J. H. Chen, N. H. Shah, S. Lachgar, P. A. Mansfield, S. Prakash, B. Green, E. Dominowska, B. Agu\"era y Arcas, N. Toma\v{s}ev, Y. Liu, R. Wong, C. Semturs, S. S. Mahdavi, J. K. Barral, D. R. Webster, G. S. Corrado, Y. Matias, S. Azizi, A. Karthikesalingam, V. Natarajan, Toward expert-level medical questio


\bibitem{li2024rt}
M. Li, H. Zhou, H. Yang, R. Zhang, RT: a Retrieving and Chain-of-Thought framework for few-shot medical named entity recognition, \textit{Journal of the American Medical Informatics Association (J. Am. Med. Inform. Assoc.)} \textbf{31}(9), 1929--1938 (2024).

\bibitem{ji2023survey}
Z. Ji, N. Lee, R. Frieske, T. Yu, D. Su, Y. Xu, E. Ishii, Y. J. Bang, A. Madotto, P. Fung, Survey of hallucination in natural language generation, \textit{ACM Computing Surveys (ACM Comput. Surv.)} \textbf{55}(12), 1--38 (2023).

\bibitem{ovadia2023fine}
O. Ovadia, M. Brief, M. Mishaeli, O. Elisha, Fine-tuning or retrieval? Comparing knowledge injection in LLMs, \textit{arXiv preprint arXiv:2312.05934} (2023).

\bibitem{lewis2020rag}
P. Lewis, E. Perez, A. Piktus, F. Petroni, V. Karpukhin, N. Goyal, H. Küttler, M. Lewis, W. Yih, T. Rocktäschel, S. Riedel, D. Kiela, Retrieval-augmented generation for knowledge-intensive NLP tasks, \textit{Advances in Neural Information Processing Systems (Adv. Neural Inf. Process. Syst.)} \textbf{33}, 9459--9474 (2020).

\bibitem{li2023understand}
M. Li, L. Huang, Understand the dynamic world: An end-to-end knowledge informed framework for open domain entity state tracking, \textit{Proceedings of the 46th International ACM SIGIR Conference on Research and Development in Information Retrieval (SIGIR)}, 842--851 (2023).

\bibitem{huang2024ritek}
J. Huang, M. Li, Z. Yao, Z. Yang, Y. Xiao, F. Ouyang, X. Li, S. Han, H. Yu, Ritek: A dataset for large language models complex reasoning over textual knowledge graphs, \textit{arXiv preprint arXiv:2410.13987} (2024).

\bibitem{zakka2024almanac}
C. Zakka, R. Shad, A. Chaurasia, A. R. Dalal, J. L. Kim, M. Moor, R. Fong, C. Phillips, K. Alexander, E. Ashley, J. Boyd, K. Boyd, K. Hirsch, C. Langlotz, R. Lee, J. Melia, J. Nelson, K. Sallam, S. Tullis, M. A. Vogelsong, J. P. Cunningham, W. Hiesinger, Almanac—retrieval-augmented language models for clinical medicine, \textit{NEJM AI} \textbf{1}(2), AIoa2300068 (2024).


\bibitem{izacard2021unsupervised}
G. Izacard, M. Caron, L. Hosseini, S. Riedel, P. Bojanowski, A. Joulin, É. Grave, Unsupervised dense information retrieval with contrastive learning, \textit{arXiv preprint arXiv:2112.09118} (2021).

\bibitem{jin2023medcpt}
Q. Jin, W. Kim, Q. Chen, D. C. Comeau, L. Yeganova, W. J. Wilbur, Z. Lu, MedCPT: Contrastive pre-trained Transformers with large-scale PubMed search logs for zero-shot biomedical information retrieval, \textit{Bioinformatics (Bioinformatics)} \textbf{39}(11), btad651 (2023).

\bibitem{touvron2023llama}
H. Touvron, T. Lavril, G. Izacard, X. Martinet, M.-A. Lachaux, T. Lacroix, B. Rozière, N. Goyal, E. Hambro, F. Azhar, A. Rodriguez, A. Joulin, É. Grave, G. Lample, LLaMA: Open and efficient foundation language models, \textit{arXiv preprint arXiv:2302.13971} (2023).

\bibitem{touvron2024llama3}
H. Touvron, T. Lavril, G. Izacard, X. Martinet, M.-A. Lachaux, T. Lacroix, B. Rozière, N. Goyal, E. Hambro, F. Azhar, A. Rodriguez, A. Joulin, É. Grave, G. Lample, A. Grattafiori, A. Dubey, A. Jauhri, A. Pandey, A. Kadian, A. Al-Dahle, A. Letman, A. Mathur, A. Schelten, A. Vaughan, A. Yang, A. Fan, A. Goyal, A. Hartshorn, A. Yang, A. Mitra, A. Sravankumar, A. Korenev, A. Hinsvark, A. Rao, A. Zhang, A. Gregerson, A. Spataru, B. Biron, B. Tang, B. Chern, C. Caucheteux, C. Nayak, C. Bi, C. Marra, C. McConnell, C. Keller, C. Touret, C. Wu, C. Wong, C. Canton Ferrer, C. Nikolaidis, D. Allonsius, D. Song, D. Pintz, D. Livshits, D. Wyatt, D. Esiobu, D. Choudhary, D. Mahajan, D. Garcia-Olano, D. Perino, D. Hupkes, E. Lakomkin, E. AlBadawy, E. Lobanova, E. Dinan, E. M. Smith, F. Radenovic, F. Guzmán, F. Zhang, G. Synnaeve, G. Lee, G. Anderson, G. Thattai, G. Nail, G. Mialon, G. Pang, G. Cucurell, H. Nguyen, H. Korevaar, H. Xu, I. Zarov, I. Arrieta Ibarra, I. Kloumann, I. Misra, I. Evtimov, J. Zhang, J. Copet, J. Lee, J. Geffert, J. Vranes, J. Park, J. Mahadeokar, J. Shah, J. van der Linde, J. Billock, J. Hong, J. Lee, J. Fu, J. Chi, J. Huang, J. Liu, J. Wang, J. Yu, J. Bitton, J. Spisak, J. Park, J. Rocca, J. Johnstun, J. Saxe, J. Jia, K. Alwala, K. Prasad, K. Upasani, K. Plawiak, K. Li, K. Heafield, K. Stone, K. El-Arini, K. Iyer, K. Malik, K. Chiu, K. Bhalla, K. Lakhotia, L. Rantala-Yeary, L. van der Maaten, L. Chen, L. Tan, L. Jenkins, L. Martin, L. Madaan, L. Malo, L. Blecher, L. Landzaat, L. de Oliveira, M. Muzzi, M. Pasupuleti, M. Singh, M. Paluri, M. Kardas, M. Tsimpoukelli, M. Oldham, M. Rita, M. Pavlova, M. Kambadur, M. Lewis, M. Si, M. Singh, M. Hassan, N. Torabi, N. Bashlykov, N. Bogoychev, N. Chatterji, N. Zhang, O. Duchenne, O. Çelebi, P. Alrassy, P. Zhang, P. Li, P. Vasic, P. Weng, P. Bhargava, P. Dubal, P. Krishnan, P. Koura, P. Xu, Q. He, Q. Dong, R. Srinivasan, R. Ganapathy, R. Calderer, R. Cabral, R. Stojnic, R. Raileanu, R. Maheswari, R. Girdhar, R. Patel, R. Sauvestre, R. Polidoro, R. Sumbaly, R. Taylor, R. Silva, R. Hou, R. Wang, S. Hosseini, S. Chennabasappa, S. Singh, S. Bell, S. Kim, S. Edunov, S. Nie, S. Narang, S. Raparthy, S. Shen, S. Wan, S. Bhosale, S. Zhang, S. Vandenhende, S. Batra, S. Whitman, S. Sootla, S. Collot, S. Gururangan, S. Borodinsky, T. Herman, T. Fowler, T. Sheasha, T. Georgiou, T. Scialom, T. Speckbacher, T. Mihaylov, T. Xiao, U. Karn, V. Goswami, V. Gupta, V. Ramanathan, V. Kerkez, V. Gonguet, V. Do, V. Vogeti, V. Albiero, V. Petrovic, W. Chu, W. Xiong, W. Fu, W. Meers, X. Martinet, X. Wang, X. Wang, X. Tan, X. Xia, X. Xie, X. Jia, X. Wang, Y. Goldschlag, Y. Gaur, Y. Babaei, Y. Wen, Y. Song, Y. Zhang, Y. Li, Y. Mao, Z. Coudert, Z. Yan, Z. Chen, Z. Papakipos, Aaditya Singh, A. Srivastava, A. Jain, A. Kelsey, A. Shajnfeld, A. Gangidi, A. Victoria, A. Goldstand, A. Menon, A. Sharma, A. Boesenberg, A. Baevski, A. Feinstein, A. Kallet, A. Sangani, A. Teo, A. Yunus, A. Lupu, A. Alvarado, A. Caples, A. Gu, A. Ho, A. Poulton, A. Ryan, A. Ramchandani, A. Dong, A. Franco, A. Goyal, A. Saraf, A. Chowdhury, A. Gabriel, A. Bharambe, A. Eisenman, A. Yazdan, B. James, B. Maurer, B. Leonhardi, B. Huang, B. Loyd, B. De Paola, B. Paranjape, B. Liu, B. Wu, B. Ni, B. Hancock, B. Wasti, B. Spence, B. Stojkovic, B. Gamido, B. Montalvo, C. Parker, C. Burton, C. Mejia, C. Liu, C. Wang, C. Kim, C. Zhou, C. Hu, C. Chu, C. Cai, C. Tindal, C. Feichtenhofer, C. Gao, D. Civin, D. Beaty, D. Kreymer, D. Li, D. Adkins, D. Xu, D. Testuggine, D. David, D. Parikh, D. Liskovich, D. Foss, D. Wang, D. Le, D. Holland, E. Dowling, E. Jamil, E. Montgomery, E. Presani, E. Hahn, E. Wood, E. Le, E. Brinkman, E. Arcaute, E. Dunbar, E. Smothers, F. Sun, F. Kreuk, F. Tian, F. Kokkinos, F. Ozgenel, F. Caggioni, F. Kanayet, F. Seide, G. Medina Florez, G. Schwarz, G. Badeer, G. Swee, G. Halpern, G. Herman, G. Sizov, G. Zhang, G. Lakshminarayanan, H. Inan, H. Shojanazeri, H. Zou, H. Wang, H. Zha, H. Habeeb, H. Rudolph, H. Suk, H. Aspegren, H. Goldman, H. Zhan, I. Damlaj, I. Molybog, I. Tufanov, I. Leontiadis, I. Veliche, I. Gat, J. Weissman, J. Geboski, J. Kohli, J. Lam, J. Asher, J. Gaya, J. Marcus, J. Tang, J. Chan, J. Zhen, J. Reizenstein, J. Teboul, J. Zhong, J. Jin, J. Yang, J. Cummings, J. Carvill, J. Shepard, J. McPhie, J. Torres, J. Ginsburg, J. Wang, K. Wu, K. U, K. Saxena, K. Khandelwal, K. Zand, K. Matosich, K. Veeraraghavan, K. Michelena, K. Li, K. Jagadeesh, K. Huang, K. Chawla, K. Huang, L. Chen, L. Garg, L. A, L. Silva, L. Bell, L. Zhang, L. Guo, L. Yu, L. Moshkovich, L. Wehrstedt, M. Khabsa, M. Avalani, M. Bhatt, M. Mankus, M. Hasson, M. Lennie, M. Reso, M. Groshev, M. Naumov, M. Lathi, M. Keneally, M. Liu, M. L. Seltzer, M. Valko, M. Restrepo, M. Patel, M. Vyatskov, M. Samvelyan, M. Clark, M. Macey, M. Wang, M. Jubert Hermoso, M. Metanat, M. Rastegari, M. Bansal, N. Santhanam, N. Parks, N. White, N. Bawa, N. Singhal, N. Egebo, N. Usunier, N. Mehta, N. Laptev, N. Dong, N. Cheng, O. Chernoguz, O. Hart, O. Salpekar, O. Kalinli, P. Kent, P. Parekh, P. Saab, P. Balaji, P. Rittner, P. Bontrager, P. Roux, P. Dollar, P. Zvyagina, P. Ratanchandani, P. Yuvraj, Q. Liang, R. Alao, R. Rodriguez, R. Ayub, R. Murthy, R. Nayani, R. Mitra, R. Parthasarathy, R. Li, R. Hogan, R. Battey, R. Wang, R. Howes, R. Rinott, S. Mehta, S. Siby, S. Bondu, S. Datta, S. Chugh, S. Hunt, S. Dhillon, S. Sidorov, S. Pan, S. Mahajan, S. Verma, S. Yamamoto, S. Ramaswamy, S. Lindsay, S. Lindsay, S. Feng, S. Lin, S. Zha, S. Patil, S. Shankar, S. Zhang, S. Zhang, S. Wang, S. Agarwal, S. Sajuyigbe, S. Chintala, S. Max, S. Chen, S. Kehoe, S. Satterfield, S. Govindaprasad, S. Gupta, S. Deng, S. Cho, S. Virk, S. Subramanian, S. Choudhury, S. Goldman, T. Remez, T. Glaser, T. Best, T. Koehler, T. Robinson, T. Li, T. Zhang, T. Matthews, T. Chou, T. Shaked, V. Vontimitta, V. Ajayi, V. Montanez, V. Mohan, V. Satish Kumar, V. Mangla, V. Ionescu, V. Poenaru, V. Mihailescu, V. Ivanov, W. Li, W. Wang, W. Jiang, W. Bouaziz, W. Constable, X. Tang, X. Wu, X. Wang, X. Wang, X. Wu, X. Gao, Y. Kleinman, Y. Chen, Y. Hu, Y. Jia, Y. Qi, Y. Li, Y. Zhang, Y. Zhang, Y. Adi, Y. Nam, Y. Wang, Y. Zhao, Y. Hao, Y. Qian, Y. Li, Z. Rait, Z. DeVito, Z. Rosnbrick, Z. Wen, Z. Yang, Z. Zhao, Z. Ma, The LLaMA 3 Herd of Models, \textit{arXiv preprint arXiv:2407.21783} (2024).


\bibitem{abdin2024phi4}
M. Abdin, J. Aneja, H. Behl, S. Bubeck, R. Eldan, S. Gunasekar, M. Harrison, R. J. Hewett, M. Javaheripi, P. Kauffmann, J. R. Lee, Y. T. Lee, Y. Li, W. Liu, C. C. T. Mendes, A. Nguyen, E. Price, G. de Rosa, O. Saarikivi, A. Salim, S. Shah, X. Wang, R. Ward, Y. Wu, D. Yu, C. Zhang, Y. Zhang, Phi-4 Technical Report, \textit{arXiv preprint arXiv:2412.08905} (2024).

\bibitem{li2025biomedrag}
M. Li, H. Kilicoglu, H. Xu, R. Zhang, BiomedRAG: A retrieval-augmented large language model for biomedicine, \textit{Journal of Biomedical Informatics} \textbf{162}, 104769 (2025).


\bibitem{yang2024qwen2}
A. Yang, B. Yang, B. Zhang, B. Hui, B. Zheng, B. Yu, C. Li, D. Liu, F. Huang, H. Wei, H. Lin, J. Yang, J. Tu, J. Zhang, J. Yang, J. Yang, J. Zhou, J. Lin, K. Dang, K. Lu, K. Bao, K. Yang, L. Yu, M. Li, M. Xue, P. Zhang, Q. Zhu, R. Men, R. Lin, T. Li, T. Tang, T. Xia, X. Ren, X. Ren, Y. Fan, Y. Su, Y. Zhang, Y. Wan, Y. Liu, Z. Cui, Z. Zhang, Z. Qiu, Qwen2.5 Technical Report, \textit{arXiv preprint arXiv:2412.15115} (2024).

\bibitem{xiong2024benchmarking}
G. Xiong, Q. Jin, Z. Lu, A. Zhang, Benchmarking retrieval-augmented generation for medicine, \textit{Findings of the Association for Computational Linguistics: ACL 2024}, 6233--6251 (2024).

\bibitem{sayers2021database}
E. W. Sayers, J. Beck, K. Brister, E. Bolton, S. Canese, D. Comeau, R. Funk, K. Kim, J. Kim, W. Klimke, W. Lee, J. Lu, T. Madden, A. Marchler-Bauer, V. Ostell, B. Phan, S. Rangwala, L. Schneider, F. Wang, D. Ye, Y. Zhang, Z. Zhao, PubMed: The NCBI database of biomedical literature, \textit{Nucleic Acids Research (Nucleic Acids Res.)} \textbf{49}(D1), D1388–D1395 (2021).


\bibitem{gurulingappa2012ade}
H. Gurulingappa, A. M. Rajput, A. Roberts, J. Fluck, M. Hofmann-Apitius, L. Toldo, Development of a benchmark corpus to support the automatic extraction of drug-related adverse effects from medical case reports, \textit{Journal of Biomedical Informatics (J. Biomed. Inform.)} \textbf{45}(5), 885--892 (2012).
\bibitem{taboureau2010chemprot}
O. Taboureau, S. K. Nielsen, K. Audouze, N. Weinhold, D. Edsgård, F. S. Roque, I. Kouskoumvekaki, A. Bora, R. Curpan, T. S. Jensen, S. Brunak, T. I. Oprea, ChemProt: a disease chemical biology database, \textit{Nucleic Acids Research (Nucleic Acids Res.)} \textbf{39}(suppl\_1), D367--D372 (2010).

\bibitem{li2023benchmarking}
M. Li, H. Zhou, R. Zhang, Benchmarking large language models in biomedical triple extraction, \textit{arXiv preprint arXiv:2310.18463} (2023).
\bibitem{zheng2021pharmkg}
S. Zheng, J. Rao, Y. Song, J. Zhang, X. Xiao, E. F. Fang, Y. Yang, Z. Niu, PharmKG: A dedicated knowledge graph benchmark for biomedical data mining, \textit{Briefings in Bioinformatics (Brief. Bioinform.)} \textbf{22}(4), bbaa344 (2021).
\bibitem{himmelstein2017hetionet}
D. S. Himmelstein, A. Lizee, C. Hessler, L. Brueggeman, S. L. Chen, D. Hadley, A. Green, P. Khankhanian, S. E. Baranzini, Systematic integration of biomedical knowledge prioritizes drugs for repurposing, \textit{eLife (eLife)} \textbf{6}, e26726 (2017).
\bibitem{gurulingappa2012ade}
H. Gurulingappa, A. M. Rajput, A. Roberts, J. Fluck, M. Hofmann-Apitius, L. Toldo, Development of a benchmark corpus to support the automatic extraction of drug-related adverse effects from medical case reports, \textit{Journal of Biomedical Informatics (J. Biomed. Inform.)} \textbf{45}(5), 885--892 (2012).
\bibitem{vasilakes2018bionli}
J. A. Vasilakes, R. Rizvi, R. Zhang, BioNLI: Generating a biomedical NLI dataset using lexico-semantic constraints for adversarial examples, University of Minnesota Digital Conservancy, \url{https://conservancy.umn.edu/handle/11299/194965} (2018).

\bibitem{pal2022medmcqa}
A. Pal, L. K. Umapathi, M. Sankarasubbu, MedMCQA: A large-scale multi-subject multi-choice dataset for medical domain question answering, \textit{Proceedings of the Conference on Health, Inference, and Learning (CHIL)}, 248--260 (2022).
\bibitem{bastan2022bionli}
M. Bastan, M. Surdeanu, N. Balasubramanian, BioNLI: Generating a biomedical NLI dataset using lexico-semantic constraints for adversarial examples, \textit{Findings of the Association for Computational Linguistics: EMNLP 2022}, 5093--5104 (2022).
\bibitem{sun2022mrc4bioer}
C. Sun, Z. Yang, L. Wang, Y. Zhang, H. Lin, J. Wang, MRC4BioER: Joint extraction of biomedical entities and relations in the machine reading comprehension framework, \textit{Journal of Biomedical Informatics (J. Biomed. Inform.)} \textbf{125}, 103956 (2022).
\bibitem{schroff2015facenet}
F. Schroff, D. Kalenichenko, J. Philbin, FaceNet: A unified embedding for face recognition and clustering, \textit{Proceedings of the IEEE Conference on Computer Vision and Pattern Recognition (CVPR)}, 815--823 (2015).

\end{thebibliography}

\newpage
\appendix
\renewcommand{\thetable}{S\arabic{table}}
\setcounter{table}{0}
\section{Supplementary Materials}

\subsection{ Paper structure summarization }
Our paper primarily focuses on evaluating the effectiveness of current Retrieval-Augmented Learning (RAL) models in the biomedical domain. The paper is organized as follows: (1) we collect and analyze existing RAL models and compile 11 biomedical datasets spanning five different tasks; (2) we propose four testbeds to systematically evaluate these models (Section Introduction); (3) we introduce methods to address key challenges, such as enhancing robustness on unlabeled data and improving negative awareness (Section Materials and Methods); and (4) we present experimental results and discussions (Section Results and Section Discussion). Additionally, we provide a glossary table~\ref{con:tasks} and ~\ref{con:Algorithms_}   summarizing the algorithms, evaluation metrics, references, descriptions, advantages, and limitations to enhance clarity and readability.

\begin{table*}[ht]
	\centering
	\renewcommand\arraystretch{1.3}
\resizebox{1\textwidth}{!}{%
	\begin{tabular} {l|l}
		\toprule 
           Task: &  triple extraction, link prediction, text classification, question
answering, and natural language inference \\
        LLMs:&   LLamA2-13B , MedLLamA-13B, LLaMA3 8B,  Phi4 14B and Qwen2.5 32B  \\
	                  % & ChemProt&  19.24 & 77.49\\
        Retrievers: &     BM25, Contriver, and MedCPT.   \\
        Dataset: &  ADE, ChemProt, GIT, PHarmKG, Hetionet, Ade-corpus-v2, SemedCLass, MedMCQA, BioNLI, DS and SDoH.  \\
    
         Metrics: &    Micro F1, Recall, Precision, Macro-F1,Negative Awareness Metrics \\
         testbed:&  Unlabeled Robustness, Counterfactual Robustness, Diverse Robustness, Negative Awarenes\\
          Our methods:&  Detect-and-Correct,contrastive
learning approach  \\
              % Sumization&   \\
        \hline
        
	\end{tabular}
 }
\vspace{+2mm}
\caption{Glossary  of tasks, LLMs, retrivers, dataset, metrics, testbeds, our method }
\label{con:tasks}
 
\end{table*}

\begin{table*}[ht]
    \centering
    \renewcommand{\arraystretch}{1.2}
    \resizebox{1\textwidth}{!}{%
    \begin{tabular}{|l|l|l|p{4.5cm}|p{4.5cm}|p{4.5cm}|}
        \hline
        \textbf{Category} & \textbf{Name} & \textbf{Reference} & \textbf{Description} & \textbf{Advantages} & \textbf{Limitations} \\
        \hline
        \multirow{3}{*}{Retrieval Algorithms} 
        & BM25 & (14) & Traditional term-based ranking model for information retrieval. & Fast and interpretable, effective for keyword-based search. & Limited to lexical overlap, does not capture semantics. \\
        \cline{2-6}
        & Contriever & (15) & Dense retrieval model using contrastive learning for embeddings. & Captures semantic meaning, works well with LLMs. & Computationally expensive, requires pretraining. \\
        \cline{2-6}
        & MedCPT &(16)& Medical domain-specific retrieval model combining contrastive pretraining and biomedical knowledge. & Tailored for biomedical data, improves retrieval quality. & Requires domain-specific data for training. \\
        \hline
        \multirow{2}{*}{Evaluation Metrics} 
        & Micro Precision, Recall, F1 & Standard ML Metric & Measures classification accuracy in terms of precision and recall. & Standard metric for classification tasks. & Does not account for retrieval-specific errors. \\
        \cline{2-6}
        & Negative Awareness Rate & Proposed in this work & Measures how well the model detects false negatives and true negatives in retrieval. & Provides insight into retrieval robustness and bias. & May conflict with task-based metrics if miscalibrated. \\
        \cline{2-6}
        \hline
        \multirow{2}{*}{Fine-Tuning \& Loss Functions} 
        & Triplet Loss & (29)& Ensures that positive instances are closer in embedding space than negative instances. & Effective for contrastive learning and retrieval tasks. & Requires careful selection of triplets for training. \\
        \cline{2-6}
        & Detect-and-Correct & This work & Identifies and corrects mislabeled instances to improve retrieval robustness. & Reduces the impact of counterfactuals and noisy retrieval. & Performance depends on retriever quality. \\
        \hline
       
    \end{tabular}}
    \caption{Glossary of Algorithms and Metrics with References, Descriptions, Advantages, and Limitations.}
    % \label{tab:glossary}
       \label{con:Algorithms_}
\end{table*}

\subsection{RESULTS OF RALS AND BACKBONE LLMS}
\label{con:results_of_llms}
\begin{table*}[ht]
	\centering
	\renewcommand\arraystretch{1.3}
\resizebox{0.8\textwidth}{!}{%
	\begin{tabular} {c|c|c|ccc|ccc|ccc|ccc}
		\toprule 
  
	\multicolumn{1}{c}  {}&\textbf{}&\multicolumn{1}{c}  {}&\multicolumn{3}{c}  {triple}& \multicolumn{3}{c}  {head entity }& \multicolumn{3}{c}  {relation} &\multicolumn{3}{c}{tail entity} \\
		%\hline
   Dataset&LLM &Approach & Precision &  Recall & F1 &  Precision &  Recall & F1 &  Precision &  Recall & F1 &  Precision &  Recall & F1 \\ 
   \hline
   \multirow{20}{*}{ADE}
    & \multirow{4}*{LLaMA2-13B}&BM25  & 30.99 &   30.88  & 30.93  &  73.96  &  73.71  & 73.84 & 94.85 & 94.54 & 94.70 & 49.10 & 48.94 & 49.02 \\
    & &   Contriever &36.07 &  36.06&36.06 & 79.76 & 79.72 & 79.74  & 93.80 & 93.76 & 93.78 & 48.99 & 48.97 & 48.98 \\
    &  &MedCPT &30.81 & 30.80 & 30.81 & 76.75 & 76.71 & 76.73 & 94.41 & 94.37 & 94.39 & 43.87 & 43.85 & 43.86\\ 
            % &random&  train& 45.12 & 45.12 & 45.12 & 83.43 & 83.43 & 83.43 & 96.10 & 96.10 & 96.10 & 54.27 & 54.27 & 54.27 \\
    &  &No Retriever& 34.94 & 34.79 & 34.86 & 83.64 & \cellcolor{green!40}\underline{83.29} & \cellcolor{green!40}\underline{83.46} & 95.29 & 94.88 & 95.08 & 42.10 & 41.92 & 42.01 \\
              % \cmidrule(lr){2-15}
              % \cmidrule(lr){2-15}
   \cmidrule(rr){2-15}
   
   &\multirow{4}*{MedLLaMA-13B}&BM25  & 33.77 & 33.76 & 33.77 & 77.06 & 77.02 & 77.04 & 94.82 & 94.77 & 94.80 & 51.20 & 51.18 &51.19 \\
    & &Contriever & 35.66 & 33.57 & 34.58 & 79.15 & 74.51 & 76.76 & 94.36 & 88.83 & 91.51 & 49.18 & 46.29 & 47.69\\
    &  & MedCPT  & 33.30 & 29.72 & 31.41 & 79.01 & 70.52 & 74.52 & 95.48 & 85.21 & 90.05 & 45.77 & 40.85 & 43.17\\ 
               % &random&train & 39.53 & 39.44 & 39.48 & 82.26 & 82.07 & 82.16 & 95.25 & 95.02 & 95.14 & 48.80 & 48.69 & 48.74 \\
    &  &No Retriever& 12.26 & 12.16 & 12.21 & 81.87 & 81.22 & 81.55 & 95.69 & \cellcolor{green!40}\underline{94.93} & \cellcolor{green!40}\underline{95.31} & 15.66 & 15.54 & 15.60 \\
    \cmidrule(rr){2-15}
    & \multirow{4}*{LLaMA3-8B}&BM25  & 27.88 & 27.70 & 27.79 & 72.87 & 72.39 & 72.63 & 94.61 & 93.99 & 94.30 & 45.79 & 45.49 & 45.64\\
    & &Contriever &34.44 & 31.13 & 32.70 & 78.91 & 71.31 & 74.92 & 93.14 & 84.18 & 88.43 & 48.57 & 43.90 & 46.12\\
    &  &MedCPT & 31.70 & 27.04 & 29.19 & 77.99 & 66.53 & 71.80 & 94.00 & 80.19 & 86.55 & 45.35 & 38.69 & 41.75\\ 
    &   &No Retriever& 9.85 & 5.49 & 7.05 & \cellcolor{green!40}\underline{83.84} & 46.76 & 60.04 &\cellcolor{green!40}\underline{ 96.30} & 53.71 & 68.96 & 13.05 & 7.28 & 9.34  \\   
    
       \cmidrule(rr){2-15}
    & \multirow{4}*{Phi4 14B}&BM25  & 37.63 & 34.93 & 36.23 & 79.16 & 73.47 & 76.21 & 94.69 & 87.89 & 91.16 & 54.43 & 50.52 & 52.40 \\
    & &   Contriever & 31.14 & 24.88 & 27.66 & 77.67 & 62.07 & 69.00 & 93.71 & 74.88 & 83.25 & 45.83 & 36.62 & 40.71  \\
    &  &MedCPT & 28.82 & 24.27 & 26.35 & 79.77 & 67.18 & 72.94 & 93.98 & 79.15 & 85.93 & 41.64 & 35.07 & 38.07 \\
            % &random&  train& 45.12 & 45.12 & 45.12 & 83.43 & 83.43 & 83.43 & 96.10 & 96.10 & 96.10 & 54.27 & 54.27 & 54.27 \\
    &  &No Retriever& 16.73 & 16.38 & 16.56 & 81.02 & 79.34 & 80.17 & 95.54 & 93.57 & 94.54 & 21.62 & 21.17 & 21.39 \\
 
          \cmidrule(rr){2-15}
    & \multirow{4}*{Qwen2.5 32B}&BM25  & 40.79 & 40.33 & 40.56 & 80.34 & 79.44 & 79.89 & 94.87 & 93.80 & 94.33 & 56.60 & 55.96 & 56.28 \\
    & &   Contriever &  \cellcolor{green!40} {\underline{46.24 }}&  \cellcolor{green!40} {\underline{45.07 }}&  \cellcolor{green!40} {\underline{45.65}} & 81.89 & 79.81 & 80.84 & 94.27 & 91.88 & 93.06 & \cellcolor{green!40}\underline{57.47} & \cellcolor{green!40}\underline{56.01 }& \cellcolor{green!40}\underline{56.73} \\
    &  &MedCPT  & 44.18 & 41.17 & 42.62 & 82.02 & 76.43 & 79.13 & 94.96 & 88.50 & 91.62 & 55.72 & 51.92 & 53.75 \\
            % &random&  train& 45.12 & 45.12 & 45.12 & 83.43 & 83.43 & 83.43 & 96.10 & 96.10 & 96.10 & 54.27 & 54.27 & 54.27 \\
    &  &No Retriever& 31.19 & 23.43 & 26.76 & 84.75 & 63.66 & 72.71 & 96.75 & 72.68 & 83.00 & 36.31 & 27.28 & 31.15  \\
       \hline
    \multirow{20}{*}{ChemProt}
     &  \multirow{4}*{LLaMA2-13B}&BM25  & 49.44 & 48.78 & 49.11 & 65.77 & 64.89 & 65.33 & 75.56 & 74.55 & 75.05 & 65.51 & 64.63 & 65.07   \\
    &&Contriever &  85.75 & 85.05 & 85.40 & 98.42 & \cellcolor{pink!40}\underline{97.61} & \cellcolor{pink!40}\underline{98.01} & 91.58 & 90.84& 91.21 & 94.02 & 93.25 & 93.63   \\
    & &MedCPT  & 86.25 & 85.40 & 85.82 &\cellcolor{pink!40}\underline{ 98.44} & 97.47 & 97.95 & 91.19 & 90.29 & 90.73 & \cellcolor{pink!40}\underline{95.53} & 94.59 & 95.06  \\ 
                % &random& train &    89.97 & 89.21 & 89.59 & 98.83 & 97.99 & 98.41 & 92.67 & 91.88 & 92.27 & 97.36 & 96.54 & 96.95\\
    & &No Retriever&   78.58 & 76.41 & 77.48 & 98.12 & 95.40 & 96.74 & 90.70 & 88.19 & 89.43 & 88.21 & 85.78 & 86.98 \\
    \cmidrule(rr){2-15}
     & \multirow{4}*{MedLLaMA-13B}&BM25 & 54.78 & 49.53 & 52.02 & 73.30 & 66.29 & 69.62 & 78.10 & 70.62 & 74.17 & 74.40 & 67.28 & 70.66 \\
    & &Contriever & 86.15 & 85.22 & 85.69 & 97.97 & 96.92 & 97.44 & 91.38 & 90.40 & 90.89 & 95.18 & 94.15 & 94.66 \\
    & & MedCPT  & 81.33 & 80.08 & 80.70 & 98.14 & 96.63 & 97.38 & 89.60 & 88.22 & 88.91 & 91.96 & 90.55 & 91.25 \\ 
    & &No Retriever&  52.10  &49.04   & 50.52  &  97.65 &  91.91 & 94.69  & 90.82  & 85.48  &88.07  &  57.91 & 54.50  &   56.15\\
                
    \cmidrule(rr){2-15}
    & \multirow{4}*{LLaMA3-8B}&BM25 & 70.23 & 69.98 & 70.10 & 83.86 & 83.57 & 83.71 & 84.30 & 84.00 & 84.15 & 91.54 & 91.22 & 91.38  \\
    &  &  Contriever &87.13 &86.68 & 86.91 & 97.92 & 97.41 & 97.67 & 91.26 & 90.78 & 91.02 & 95.56 & \cellcolor{pink!40}\underline{95.06} & \cellcolor{pink!40}\underline{95.30} \\
    & &MedCPT  & 86.62 & 83.07 & 84.81 & 98.06 & 94.04 & 96.01 &\cellcolor{pink!40}\underline{ 92.54} & 88.74 & 90.60 & 94.06 & 90.20 & 92.09 \\
    & &No Retriever& 23.21 & 19.72 & 21.32 & 98.63 & 83.80 & 90.61 & 91.75 & 77.95 & 84.29 & 26.09 & 22.16 & 23.97  \\
     \cmidrule(rr){2-15}
    & \multirow{4}*{Phi4 14B}&BM25  & 82.46 & 80.10 & 81.26 & 95.09 & 92.38 & 93.71 & 90.42 & 87.84 & 89.11 & 91.95 & 89.33 & 90.62\\
    & &   Contriever& 71.36 & 68.12 & 69.70 & 98.60 & 94.12 & 96.31 & 93.39 & 89.15 & 91.22 & 76.66 & 73.18 & 74.88 \\
    &  &MedCPT & 81.48 & 77.57 & 79.48 & 97.71 & 93.02 & 95.31 & 92.09 & 87.67 & 89.82 & 88.45 & 84.21 & 86.28   \\
            % &random&  train& 45.12 & 45.12 & 45.12 & 83.43 & 83.43 & 83.43 & 96.10 & 96.10 & 96.10 & 54.27 & 54.27 & 54.27 \\
    &  &No Retriever& 79.09 & 75.45 & 77.23 & 99.02 & 94.47 & 96.70 & 93.38 & 89.09 & 91.19 & 85.18 & 81.27 & 83.18\\
 
          \cmidrule(rr){2-15}
    & \multirow{4}*{Qwen2.5 32B}&BM25  & 86.46 & 86.01 & 86.24 & 98.27 & 97.76 & 98.02 & 90.18 & 89.70 & 89.94 & 95.94 & 95.43 & 95.68\\
    & &   Contriever & 87.26 & 87.06 & 87.16 & 98.34 & 98.11 & 98.22 & 90.06 & 89.85 & 89.95 & 97.20 & 96.97 & 97.09\\
    &  &MedCPT & 87.86 & 87.58 & 87.72 & 98.51 & 98.20 & 98.35 & 90.90 & 90.61 & 90.75 & 96.38 & 96.07 & 96.23\\
            % &random&  train& 45.12 & 45.12 & 45.12 & 83.43 & 83.43 & 83.43 & 96.10 & 96.10 & 96.10 & 54.27 & 54.27 & 54.27 \\
    &  &No Retriever& \cellcolor{pink!40} \underline{88.38} &\cellcolor{pink!40} \underline{ 87.81} &\cellcolor{pink!40} \underline{ 88.09 }& 98.68 & 98.05 & 98.37 &   91.92 & \cellcolor{pink!40} \underline{ 91.33 }& \cellcolor{pink!40} \underline{ 91.63} & 95.49 & 94.88 & 95.19\\
     
    \hline
    \multirow{20}{*}{GIT} & \multirow{4}*{LLaMA2-13B}&BM25  & 60.81 & 54.73 & 57.61 & 76.94 & 69.25 & 72.89 & 76.70 & 69.03 & 72.67 & 77.78 & 70.00 & 73.68 \\
    & &Contriever & 74.78 & 72.37 & 73.55 & 89.44 & 86.56 & 87.98 & \cellcolor{blue!20}\underline{83.22} & 80.54 & 81.86 & 87.11 & 84.30 & 85.68\\
    &  &MedCPT  & 75.64 & 73.44 & 74.52 & \cellcolor{blue!20}\underline{92.03 }& \cellcolor{blue!20}\underline{89.35} & \cellcolor{blue!20}\underline{90.67} & 83.06 & 80.65 & 81.83 & 89.04 & 86.45 & 87.73 \\ 
             % &random&train&72.14  &  70.75   & 71.44  &  88.37  & 86.66    & 87.51    & 81.57   & 80.00   &80.78    &  86.62  & 84.94&85.77  \\
    &   &No Retriever&  61.76 & 56.45 & 58.99 & 84.12 & 76.88 & 80.34 & 73.53 & 67.20 & 70.22 & 82.94 & 75.81 & 79.21 \\
    \cmidrule(rr){2-15}
    &  \multirow{4}*{MedLLaMA-13B}&BM25 &58.59 & 57.20 & 57.89 & 79.30 & 77.42 & 78.35 & 72.91 & 71.18 & 72.03 & 78.63 & 76.77 & 77.69 \\
    & &Contriever & 65.95 & 65.81 & 65.88 & 85.34 & 85.16 & 85.25 & 75.86 & 75.70 & 75.78 & 83.84 & 83.66 & 83.75 \\
   &  &  MedCPT  & \cellcolor{blue!20} \underline{75.65} &\cellcolor{blue!20}  \underline{74.84} & \cellcolor{blue!20} \underline{75.24} & 90.22 & 89.25 & 89.73 & 82.39 & 81.51 & 81.95 & \cellcolor{blue!20}\underline{90.00 }&\cellcolor{blue!20} \underline{89.03 }&\cellcolor{blue!20} \underline{89.51}  \\ 
              % &random& train&58.70  &36.98     & 45.38  &  90.44  &56.99     & 69.92    & 83.27   & 52.47   &   64.37 & 69.28   & 43.65&53.56  \\
    &   &No Retriever&  42.60 & 41.51 & 42.05 & 89.18 & 86.88 & 88.02 & 75.28 & 73.33 & 74.29 & 52.76 & 51.40 & 52.07 \\ 
    \cmidrule(rr){2-15}
    &  \multirow{4}*{LLaMA3-8B}&BM25 &62.72 & 62.58 & 62.65 & 79.31 & 79.14 & 79.22 & 79.74 & 79.57 & 79.66 & 77.80 & 77.63 & 77.72  \\
    &   &Contriever &74.76 & 67.53 & 70.96 & 89.76 & 81.08 & 85.20 & 82.14 & 74.19 & 77.97 & 88.33 & 79.78 & 83.84\\
    & &MedCPT & 66.20 & 50.97 & 57.59 & 87.15 & 67.10 & 75.82 & 82.96 & 63.87 & 72.17 & 78.77 & 60.65 & 68.53 \\ 
              % &random&train  &  75.48 & 75.48 & 75.48 & 87.63 & 87.63 & 87.63 & 82.80 & 82.80 & 82.80 & 87.63 & 87.63 & 87.63  \\
     & &No Retriever&  75.81 & 72.80 & 74.27 & 87.79 & 84.30 & 86.01 & 85.55 &\cellcolor{blue!20} \underline{82.15} & \cellcolor{blue!20}\underline{83.82} & 86.67 & 83.23 & 84.91 \\
      \cmidrule(rr){2-15}
    & \multirow{4}*{Phi4 14B}&BM25 & 64.18 & 57.42 & 60.61 & 84.62 & 75.70 & 79.91 & 78.37 & 70.11 & 74.01 & 77.16 & 69.03 & 72.87  \\
    & &   Contriever & 75.49 & 74.19 & 74.84 & 89.50 & 87.96 & 88.72 & 83.37 & 81.94 & 82.65 & 87.75 & 86.24 & 86.98\\
    &  &MedCPT  & 73.51 & 71.61 & 72.55 & 88.30 & 86.02 & 87.15 & 81.68 & 79.57 & 80.61 & 87.64 & 85.38 & 86.49\\
            % &random&  train& 45.12 & 45.12 & 45.12 & 83.43 & 83.43 & 83.43 & 96.10 & 96.10 & 96.10 & 54.27 & 54.27 & 54.27 \\
    &  &No Retriever& 30.28 & 29.89 & 30.09 & 87.15 & 86.02 & 86.58 & 76.69 & 75.70 & 76.19 & 41.83 & 41.29 & 41.56\\
 
          \cmidrule(rr){2-15}
    & \multirow{4}*{Qwen2.5 32B}&BM25 & 66.19 & 64.84 & 65.51 & 85.62 & 83.87 & 84.74 & 78.49 & 76.88 & 77.68 & 78.59 & 76.99 & 77.78 \\
    & &   Contriever & 73.12 & 73.12 & 73.12 & 87.20 & 87.20 & 87.20 & 81.29 & 81.29 & 81.29 & 85.70 & 85.70 & 85.70\\
    &  &MedCPT & 69.46 & 69.46 & 69.46 & 86.24 & 86.24 & 86.24 & 79.35 & 79.35 & 79.35 & 84.95 & 84.95 & 84.95\\
            % &random&  train& 45.12 & 45.12 & 45.12 & 83.43 & 83.43 & 83.43 & 96.10 & 96.10 & 96.10 & 54.27 & 54.27 & 54.27 \\
    &  &No Retriever& 68.72 & 68.28 & 68.50 & 86.36 & 85.81 & 86.08 & 78.79 & 78.28 & 78.53 & 84.31 & 83.76 & 84.03\\
           \bottomrule
	\end{tabular}
 }
\vspace{+2mm}
\caption{Results of various approaches for triple extraction on ADE, ChemProt, and GIT. Underline with shade (green, pink, and blue) indicates the best performance on ADE, ChemProt, and GIT separately. }
\label{con:full_results_1}
 
\end{table*}

\begin{table*}[ht]
	\centering
	\renewcommand\arraystretch{1.3}
\resizebox{0.8\textwidth}{!}{%
\Huge
	\begin{tabular} {l|c|ccc|ccc|ccc|ccc|ccc|c}
		\toprule 
    &\multicolumn{1}{c}  {}&\multicolumn{6}{c}  {Link Prediction}&\multicolumn{6}{c}  {Text Classification}&\multicolumn{3}{c}  {Question Answering}&\multicolumn{1}{c}  {Natural Language Inference}\\ 
      \cmidrule(rr){3-8}\cmidrule(rr){9-14}\cmidrule(rr){15-17} \cmidrule(rr){18-18}
		&\multicolumn{1}{c}  {}&\multicolumn{3}{c}{PHarmKG}& \multicolumn{3}{c}{Hetionet}& \multicolumn{3}{c}{Ade-corpus-v2}& \multicolumn{3}{c}{SemClass}  & \multicolumn{3}{c}{MedMCQA} &\multicolumn{1}{c}{BioNLI} \\
		%\hline
		  LLM&Approach & Precision &  Recall & F1 &  Precision &  Recall & F1 & Precision &  Recall & F1 &  Precision &  Recall & F1 & Precision &  Recall & F1&  Macro-avg F1 \\ 	
   
	      \midrule
              \multirow{4}*{LLaMA2-13B}&BM25  & 97.60   & 97.60      &97.60     & 82.37   & 82.37    &82.37  & 95.40  &95.40      & 95.40   &  75.50  &   75.50   &  75.50   &  40.38  &  40.49   & 40.42 & 45.10  \\
            &Contriever &  \cellcolor{green!40} \underline{ 98.00}  &  \cellcolor{green!40}  \underline{ 98.00}    &  \cellcolor{green!40} \underline{98.00 } &   77.00& 77.00    & 77.00& 96.60   & 96.60     &  96.60  &  \cellcolor{green!40}  \underline{79.33}   &     \cellcolor{green!40}  \underline{79.33 } &    \cellcolor{green!40} \underline{79.33}   &  35.53  &  35.52   &  35.52  &  35.12  \\
            & MedCPT &   97.40  &  97.40    &  97.40&  81.60  &  81.60   &  81.60 & \cellcolor{green!40}  \underline{96.80 }    &   \cellcolor{green!40} \underline{96.80}   & \cellcolor{green!40}  \underline{96.80}    & 78.50    &  78.50    &     78.50&  36.78  &  36.93   &  36.80  &   69.21 \\
             % &random &  train &testing  &     &   & 79.88   &  79.88   & 79.88   \\
              &No Retriever &  97.60  &   97.60  & 97.60  &  80.80  &  80.80   &  80.80 & 96.40  & 96.40    &  96.40 &   77.66  & 77.66   & 77.66& 41.63   & 41.52  & 41.52 &  62.62 \\
            \hline
               \multirow{4}*{MedLLaMA-13B}&BM25 & 95.00   &   95.00    &    95.00 &  \cellcolor{green!40}  \underline{90.40 } & \cellcolor{green!40} \underline{ 90.40 }   &   \cellcolor{green!40} \underline{ 90.40}   & 95.60   & 95.60    &  95.60 &  72.67  &   72.67  &   72.67 &  37.81  & 37.96    &  37.86 &    48.81\\
            &Contriever &  97.00  &    97.00  &   97.00  &   77.20 &  77.20   &   77.20  &  95.60  &  95.60   &  95.60 &  77.66 & 77.66   & 77.66&  29.82  &   29.75  &  29.77&      53.07  \\
            & MedCPT&     97.40    &   97.40     &  97.40      &   84.40 &   84.40  &   84.40 & 95.40 & 95.40    & 95.40  &76.16    &  76.16    &   76.16&  33.86  &  34.04   &  33.88 &    53.68 \\
             % &random &  train &  running &     &   &    &     &    \\
               &No Retriever &  97.20  &  97.20    &97.20   &  78.54  &  78.54   &  78.54 & 95.40  &  95.40   & 95.40  &   64.00 &    64.00 &  64.00  &  46.79  &  46.41   & 46.47 &    61.07 \\
             \hline
        \multirow{4}*{LLaMA3-8B}&BM25   & 96.80  & 96.80    &  96.80 &  81.80  &   81.80  &   81.80&94.80   & 94.80    & 94.80   & 75.50  &   75.50    &  75.50  &  37.73  &  38.90   & 37.79 & 19.17 \\
            &Contriever &  96.60   &   96.60    &  96.60  &  73.40  &  73.40   &   73.40& 94.60  &  94.60    & 94.60   &   75.83 &  75.83 &75.83&  28.11  &  28.12   &  28.11 & 63.85 \\
            & MedCPT  &  97.00  &  97.00   & 97.00   & 83.00   &   83.00  &  83.00 & 95.40 & 95.40    & 95.40  & 74.67 &  74.67   & 74.67&  31.57  &  31.82   &  31.56 &  56.89   \\
             % &random &  train & testing   &     &   & 80.29   &  80.29    & 80.29    \\
               &No Retriever &  97.20 & 97.20    &  97.20 &   81.80 &  81.80   &   81.80&93.80   & 93.80      & 93.80    &   73.16 &   73.16   &  73.16  &   56.93 & 55.43    &  55.91 & 6.71 \\
          
        \hline
        \multirow{4}*{Phi4 14B}&BM25    &    97.20 &   97.20    & 97.20 & 88.80   & 88.80     &  88.80 & 88.20  &  88.20     &  88.20   & 74.67   &  74.67   & 74.67   &  39.33  &  39.62   &  39.42 & 67.65   \\
            &Contriever  &  31.60  &   31.60   & 31.60 &  74.20  &   74.20  &  74.20 & 90.80  &  90.80     &  90.80   & 75.17   &  75.17   &  75.17  &  35.38  &  34.81   & 33.89  & 56.65  \\
            & MedCPT   &  96.80  &   96.80   & 96.80 &  82.80  &  82.80   &  82.80 & 90.60   & 90.60      & 90.60    &  75.00  & 75.00     &  75.00   &  29.72  &  26.74   & 27.72  &  85.45 \\
               &No Retriever &  97.60  &  97.60   &  97.60&   82.20 &  82.20  & 82.20  & 89.20  &   89.20    & 89.20    &  72.50  & 72.50    & 72.50   & \cellcolor{green!40} \underline{66.46} &  \cellcolor{green!40} \underline{66.01}  & \cellcolor{green!40} \underline{66.18}  & 82.50  \\
                \bottomrule
      \multirow{4}*{Qwen2.5 32B}&BM25   &   97.20 & 97.20     & 97.20 &  82.00  & 82.00    &  82.00 & 94.20  & 94.20      & 94.20    &  78.50  &  78.50   & 78.50   & 29.67  &  27.41   &  28.39 & 81.29  \\
            &Contriever  &  97.60&  97.60    & 97.60 & 75.80   &  75.80   & 75.80  &  94.20 &  94.20    &  94.20   &  76.33  & 76.33    &  76.33  &  28.96  &  27.44   & 28.10  &  89.99 \\
            & MedCPT   & 96.20   &  96.20    & 96.20&  82.80  & 82.80    & 82.80  & 95.00 &    95.00   & 95.00    & 76.83   & 76.83   & 76.83  & 29.15  &  27.53   & 28.22 & \cellcolor{green!40} \underline{91.19}  \\
               &No Retriever &  90.40  &  90.40    & 90.40 &  81.00  & 81.00    &  81.00 & 94.80  &   94.80    &  94.80    & 48.00   & 48.00    & 48.00   &  24.69  &  18.30   &  20.91  &  63.20 \\
                \bottomrule
	\end{tabular}
 }
\vspace{+2mm}
\caption{Results of various approaches for link prediction, text classification, question answering, and natural language inference.  Underline with green shade indicates the best performance on each dataset.}
\label{con:full_results_2}
 
\end{table*}

\begin{table*}[ht]
	\centering
	\renewcommand\arraystretch{1.3}
\resizebox{0.8\textwidth}{!}{%
	\begin{tabular} {c|c|ccc|ccc}
		\toprule 
  
	\textbf{}&\multicolumn{1}{c}  {}&\multicolumn{3}{c}  {DS link prediction}& \multicolumn{3}{c}  {SDH classification } \\
		%\hline
  LLM &Approach & Precision &  Recall & F1 &  Precision &  Recall & F1 \\ 
   \hline
     \multirow{4}*{LLaMA2-13B}&BM25  & 75.86 &       75.86& 75.86    & 63.71     &   63.71   &  63.71 \\
     &   Contriever  &76.50 &  76.50    &  76.50  & 61.18    &    61.18 &  61.18\\
      &MedCPT   &78.01  & 78.01     & 78.01   &  68.77    &  68.77  & 68.77\\
      &No Retriever &77.15  &77.15     &77.15   &    72.57 &   72.57 &72.57 \\
      \hline
       \multirow{4}*{MedLLaMA-13B}&BM25 &71.33 &     71.33  & 71.33    &  65.40    & 65.40    & 65.40 \\
     &   Contriever  & 75.64&  75.64    &  75.64  &  56.96    &  56.96      &    56.96 \\
      &MedCPT  &72.84&72.84      &72.84    &  66.66   & 66.66   &66.66 \\
      &No Retriever &80.38  &80.38      & 80.38   & 67.71    &67.71     &67.71  \\
      \hline
       \multirow{4}*{LLaMA3-8B}&BM25 &75.43 &   75.43   &  75.43  &     71.30   &    71.30  & 71.30  \\
     &   Contriever  & 78.01& 78.01     &  78.01  &    72.57   & 72.57   & 72.57  \\
      &MedCPT  & 76.29 &   76.29   &76.29    &     71.72  &      71.72&  71.72 \\
      % &No Retriever &80.38 &   80.38    &  80.38   &   85.65  &  85.65  &85.65 \\
       &No Retriever &80.38 &   80.38    &  80.38   &   57.00  & 57.00 &57.00 \\
      \hline
       \multirow{4}*{Phi4 14B}&BM25  & 60.75 &  60.75     &   60.75  &   78.90  &78.90     &78.90  \\
     &   Contriever  &53.58  &53.58       &  53.58   &   81.43   &   81.43  & 81.43 \\
      &MedCPT   & 57.38  &  57.38     &  57.38   &    80.16  &  80.16  & 80.16\\
      % &No Retriever & 54.05 &     54.05 & 54.05   &   88.18   &88.18     & 88.18 \\
       &No Retriever & 54.05 &     54.05 & 54.05   &   74.68    &74.68      & 74.68   \\
      \hline
       \multirow{4}*{Qwen2.5 32B}&BM25 &72.44  &    72.44  &  72.44  &  78.05    &   78.05  & 78.05 \\
     &   Contriever  &77.37 & 77.37     & 77.37   &    80.16  &     80.16 &  80.16 \\
      &MedCPT & 71.78 & 71.78     &71.78    &    72.99  & 72.99   &72.99 \\
      &No Retriever & 76.93 & 76.93  & 76.93 &   74.26   &  74.26  &74.26   \\
       % &No Retriever & 76.93 & 76.93  & 76.93 &   83.96   &   83.96   & 83.96  \\
      \hline
	\end{tabular}
 }
\vspace{+2mm}
\caption{Results of various approaches for two private datasets DS and SDH. }
\label{con:full_results_2}
 
\end{table*}

\end{document}